\documentclass[3p,10pt,number]{elsarticle}

\usepackage{amssymb}
\usepackage{amsmath}

\journal{An Elsevier Journal}
\usepackage{url}

\usepackage{breakurl}
\begin{document}

\begin{frontmatter}



\title{Semantic Reasoning in Medicine: The Role of Knowledge Graphs Across Five Key Domains} 



\author[aff1]{Haniye Sherafatmandjoo}
\ead{sherafatmandhaniye@aut.ac.ir}
\author[aff1]{Mohammad Akbari\corref{cor1}}
\ead{akbari.ma@gmail.com }
\author[aff1]{Zahed Rahmati\corref{cor1}} 
\ead{zrahmati@aut.ac.ir }
\cortext[cor1]{Corresponding author}
\affiliation[aff1]{organization={Department of Mathematics and Computer Science },
            addressline={Amirkabir University of Technology}, 
            city={Tehran},
            country={Iran}}
\vspace{-10cm}
\begin{abstract}
Knowledge graphs (KGs) have emerged as a promising solution for integrating and reasoning over complex biomedical and clinical data in healthcare. By representing structured relationships among entities such as diseases, drugs, symptoms, and patient records, KGs provide a semantic backbone for decision-making, prediction, recommendation, and personalized care. Recent advances have demonstrated their utility across diverse medical applications—including clinical decision support systems, disease and treatment outcome prediction, health recommender systems, precision medicine, and medical question answering—where KGs often enhance interpretability, semantic coherence, and patient-specific reasoning. In parallel, a growing body of work focuses on medical KG generation itself, proposing frameworks that construct graphs from EHRs, clinical narratives, biomedical literature, and web resources using ontologies, semantic web technologies, deep–learning–based information extraction, and hybrid neuro-symbolic pipelines. Despite this progress, significant challenges remain, including limited and fragmented knowledge coverage, difficulties in aligning heterogeneous data sources, the fragility of current reasoning and representation-learning methods on dense multi-relational graphs, and unresolved issues related to privacy, bias, and accountability. This survey reviews and categorizes current research on KGs in medicine along both application-oriented and methodology-oriented dimensions, discusses their benefits and technical foundations, and outlines key limitations and open research directions. By analyzing trends, architectures, and evaluation practices, this work aims to guide future developments in KG-driven medical AI systems and support their safe and effective integration into healthcare environments.
\end{abstract}



\begin{keyword}
Medical Knowledge Graphs \sep Clinical Decision Support \sep Disease Prediction \sep Health Recommender Systems \sep Precision Medicine \sep Question Answering \sep Knowledge Graph Generation \sep Data Integration in Healthcare


\end{keyword}

\end{frontmatter}



\section{Introduction}
\label{sec:sec1}

The exponential growth of healthcare data has created an urgent need for stable, coherent, and semantically rich structures to store and analyze information. Among the available representation paradigms, knowledge graphs (KGs) have emerged as a powerful way to organize complex, heterogeneous, and often sensitive medical data. Initially popularized by Google in 2012 to enhance web search, knowledge graphs are now widely used across domains, including biomedicine and healthcare. Conceptually, a KG consists of entities (nodes) and relations (edges), often encoded as directed triples of the form (head entity – relation – tail entity), such as (Drug – interacts with – Drug) or (Gene – translates to – Protein). These graphs are typically grounded in one or more ontologies that define the types of entities and relations, thereby ensuring semantic consistency across data sources.
Medical data, however, present unique challenges. They are inherently complex, multi-scale (from molecular measurements to clinical outcomes), and frequently distributed across heterogeneous systems such as hospital information systems, registries, and research databases. Questionnaire-based data may be noisy or unstable, further complicating integration and modeling. In this context, Electronic Health Records (EHRs) provide a relatively structured and consistent source of clinical information. EHRs capture digitized patient data—including diagnoses, procedures, medications, laboratory results, and clinical notes—which can be mapped to ontologies and represented in KG form to support downstream analytical tasks \cite{lu2023disease}.

A growing body of work demonstrates that knowledge graphs offer an effective scaffold for representing and integrating biomedical and clinical information. Researchers either construct task-specific KGs or leverage large, publicly available graphs. For instance, PrimeKG is a multimodal KG tailored to precision medicine that integrates data from 20 curated sources—including drug, genetic, biological, and clinical databases—to unify dispersed biomedical knowledge across domains \cite{chandak2023building}. With more than 129,000 nodes and 4 million edges, PrimeKG connects diseases, proteins, drugs, biological processes, and anatomical structures, and augments this structure with clinical narratives and textual descriptions. This combination of structured relations and rich semantic context is designed to support AI-driven tasks such as diagnosis, risk stratification, and drug repurposing.
Other large-scale efforts similarly illustrate how KGs can consolidate heterogeneous biomedical resources into a coherent whole. \cite{quan2023aimedgraph} proposed a multi-relational KG for precision medicine that links genes, variants, diseases, drugs, clinical trials, and supporting studies, while explicitly modeling relation types such as variant–disease effects, drug interactions, and adverse drug reactions. Relations are annotated with evidence levels and confidence scores, enabling downstream reasoning over both explicit and implicit knowledge. SPOKE is another example of a broad biomedical KG, integrating information from 41 databases and 11 ontologies to connect millions of entities spanning genetics, biochemistry, pharmacology, and disease phenotypes \cite{morris2023scalable}. In SPOKE, ontologies play a central role in harmonizing entity names and relations, thereby supporting interoperability and complex query capabilities across diverse data sources. Additionally \cite{su2022interactive} introduced an interactive knowledge graph–based platform designed to support clinical research related to COVID-19. During the pandemic, a vast number of studies, clinical reports, and drug response data were published at unprecedented speed and volume, creating significant challenges for researchers in retrieving and synthesizing useful information. To address this, the authors proposed a framework that combines Named Entity Recognition (NER) with text summarization techniques to automatically extract key information regarding treatments, side effects, and patient demographics in a structured format. The framework incorporates tools such as GROBID for structured extraction from PDF documents, spaCy for preprocessing and text summarization, and advanced NER models like Stanza to accurately detect clinical entities. After extracting the relevant entities from text, syntactic dependency analysis is used to identify relationships among them, and rule-based methods help formalize these connections. The resulting platform enables visual exploration and interaction with the knowledge graph, allowing researchers to browse, search, and analyze medical information through a graphical interface by interacting with nodes and edges, diving deep into clinical evidence, treatment outcomes, and related scientific literature.

Beyond these large, generic resources, many domain-focused KGs have been developed to capture knowledge in specialized clinical or biological subfields. For example, MiKG concentrates on the gut–brain axis and the role of the microbiome in mental health \cite{liu2021exploring}. In that work, the authors systematically gathered and ranked evidence from the literature, extracted entities such as microbiota species, neurotransmitters, and psychiatric disorders, and aligned the resulting triples with standard biomedical ontologies (e.g., UMLS, MeSH, SNOMED CT, KEGG). The outcome is an interoperable KG that structures fragmented knowledge into a form amenable to computational analysis. Similarly, PharmKG focuses on pharmacological relationships among genes, drugs, and diseases; it integrates curated resources like DrugBank, PharmGKB, OMIM, TTD, and SIDER, and enriches entities with multi-omics features and textual representations \cite{zheng2021pharmkg}. These examples illustrate a general trend: medical KGs increasingly combine curated ontological structure with high-dimensional features and textual information, enabling richer modeling of biomedical phenomena.
From a methodological perspective, KGs provide a flexible backbone that can be exploited by a range of reasoning and learning techniques. Traditional rule-based approaches (e.g., if–then rules) and classical machine learning algorithms (such as logistic regression, random forests, or support vector machines) can operate on features derived from KGs, supporting tasks like risk prediction or alert generation. More recently, graph representation learning—especially graph neural networks (GNNs)—has become a central paradigm for modeling medical KGs. GNNs iteratively aggregate information from neighboring nodes to learn context-aware representations that capture multi-hop relationships and graph structure, making them well suited for complex tasks such as disease risk prediction, drug–drug interaction modeling, and patient stratification \cite{paul2024systematic}. These models can integrate both domain knowledge encoded in the KG and personalized patient data from EHRs, thereby bridging population-level knowledge with individual-level information.

In parallel, the rapid progress of large language models (LLMs) has opened new opportunities to combine structured knowledge graphs with powerful text-based reasoning systems. KGs can provide factual grounding and explicit structure to mitigate hallucinations and improve explainability, while LLMs can serve as natural language interfaces that interpret user queries and verbalize complex graph-derived insights. Such KG–LLM integration holds particular promise for applications like clinical decision support, question answering, and interactive exploration of medical knowledge, where both accuracy and interpretability are critical.
Against this background, there is a need for a focused synthesis of how knowledge graphs are being used across key healthcare application areas. Existing surveys have reviewed graph-based methods or specific modeling techniques in medicine (e.g., GNNs in healthcare), but a cross-cutting view that organizes KG-based applications by clinical task and highlights their shared strengths and limitations remains limited. This survey addresses that gap by providing a structured overview of recent research on medical knowledge graphs, with an emphasis on their role in end-to-end healthcare applications.
Specifically, this work concentrates on studies that employ KGs—explicitly or implicitly—as core components in:
(i) clinical decision support systems,
(ii) disease and treatment outcome prediction,
(iii) health recommender systems,
(iv) precision medicine, and
(v) medical question answering.

In addition to organizing KG-based applications by clinical task, this review also synthesizes recent methodological work on medical knowledge graph generation, highlighting common design patterns and open challenges in constructing high-quality, domain-specific graphs for healthcare. The included studies were selected based on recency, scholarly relevance, diversity of clinical applications, and the centrality of KGs within the proposed frameworks. By organizing the literature according to these five task categories, we aim to (1) clarify how medical KGs are constructed and leveraged in different clinical contexts, (2) compare methodological patterns and design choices across domains, (3) identify common challenges and gaps—such as data quality, interoperability, evaluation, and explainability—and (4) outline promising directions for future research at the intersection of knowledge graphs, machine learning, and healthcare.

\section{Taxonomy of Knowledge Graph Applications in Healthcare}
\label{sec:sec2}
Given the wide-ranging applications of knowledge graphs in the medical domain, a conceptual framework is essential for organizing and analyzing the existing literature in a coherent and comparative manner. In this survey, we propose a taxonomy that classifies the use of knowledge graphs into five key application domains: (1) Clinical Decision Support Systems (CDSS), (2) Predictive Modeling of Diseases and Treatment Outcomes, (3) Health Recommender Systems, (4) Precision Medicine, and (5) Medical Question Answering Systems. An overview of these five application domains and the cross-cutting themes of explainability, hybrid AI, and personalization is shown in Figure1.

\begin{figure}[htbp]
\centering
\includegraphics[width=.7\linewidth]{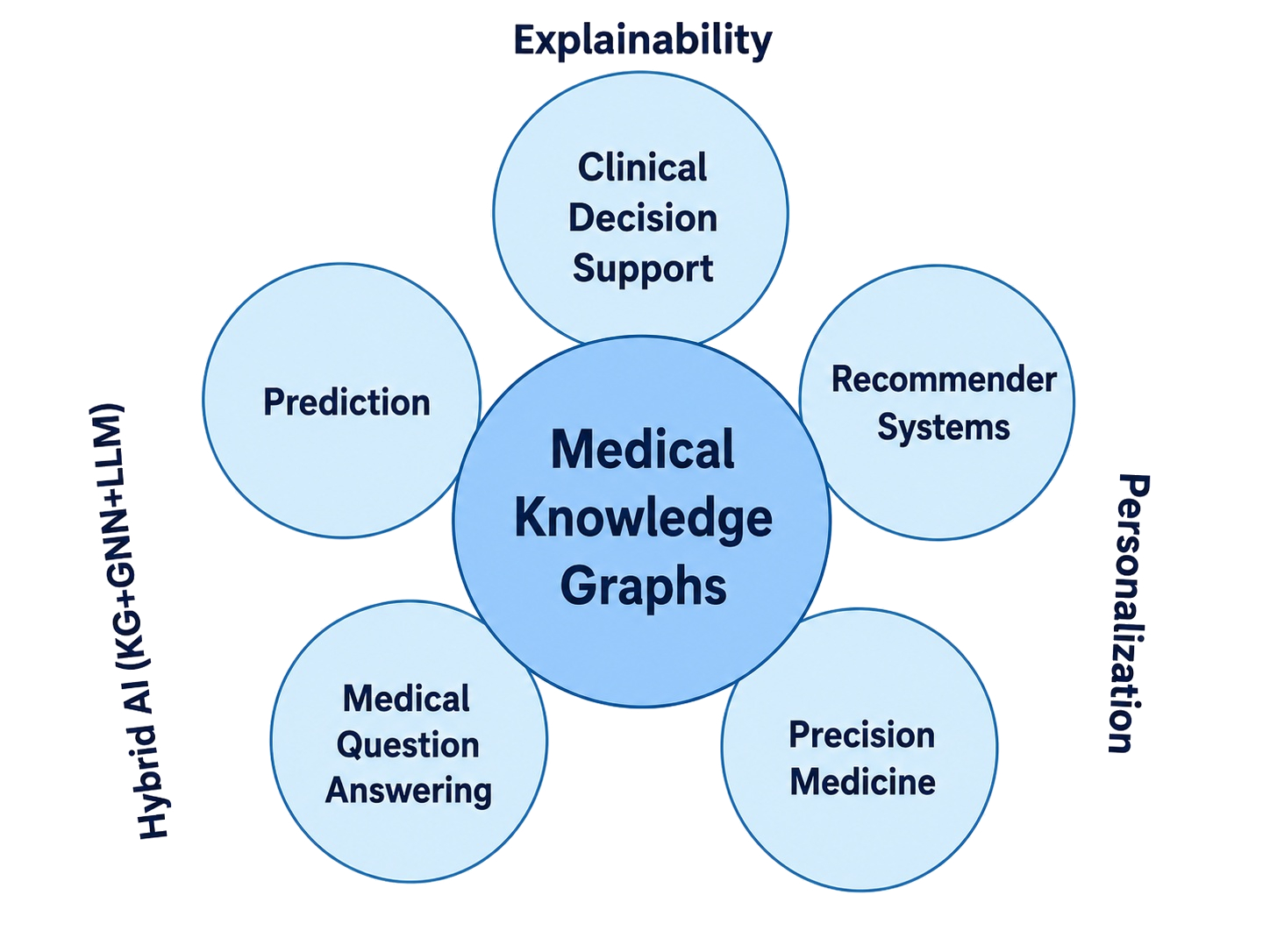}
\caption{Conceptual overview of the five application domains covered in this survey—clinical decision support, prediction, recommender systems, precision medicine, and medical question answering. Medical knowledge graphs form the semantic core that supports these tasks, while explainability, hybrid AI (KG + GNN + LLM), and personalization act as cross-cutting themes that shape how KG-based systems are designed and evaluated.}\label{fig:fig1}
\end{figure}
Each of these categories features distinct goals, data types, technical methodologies, and domain-specific challenges. This taxonomy enables a more structured analysis and facilitates a deeper understanding of ongoing trends and research directions in the field. An overview of these five application domains and the representative KG-based tasks reviewed in each category is shown in Figure2.

\begin{figure*}[t]
\centering
\includegraphics[width=\textwidth]{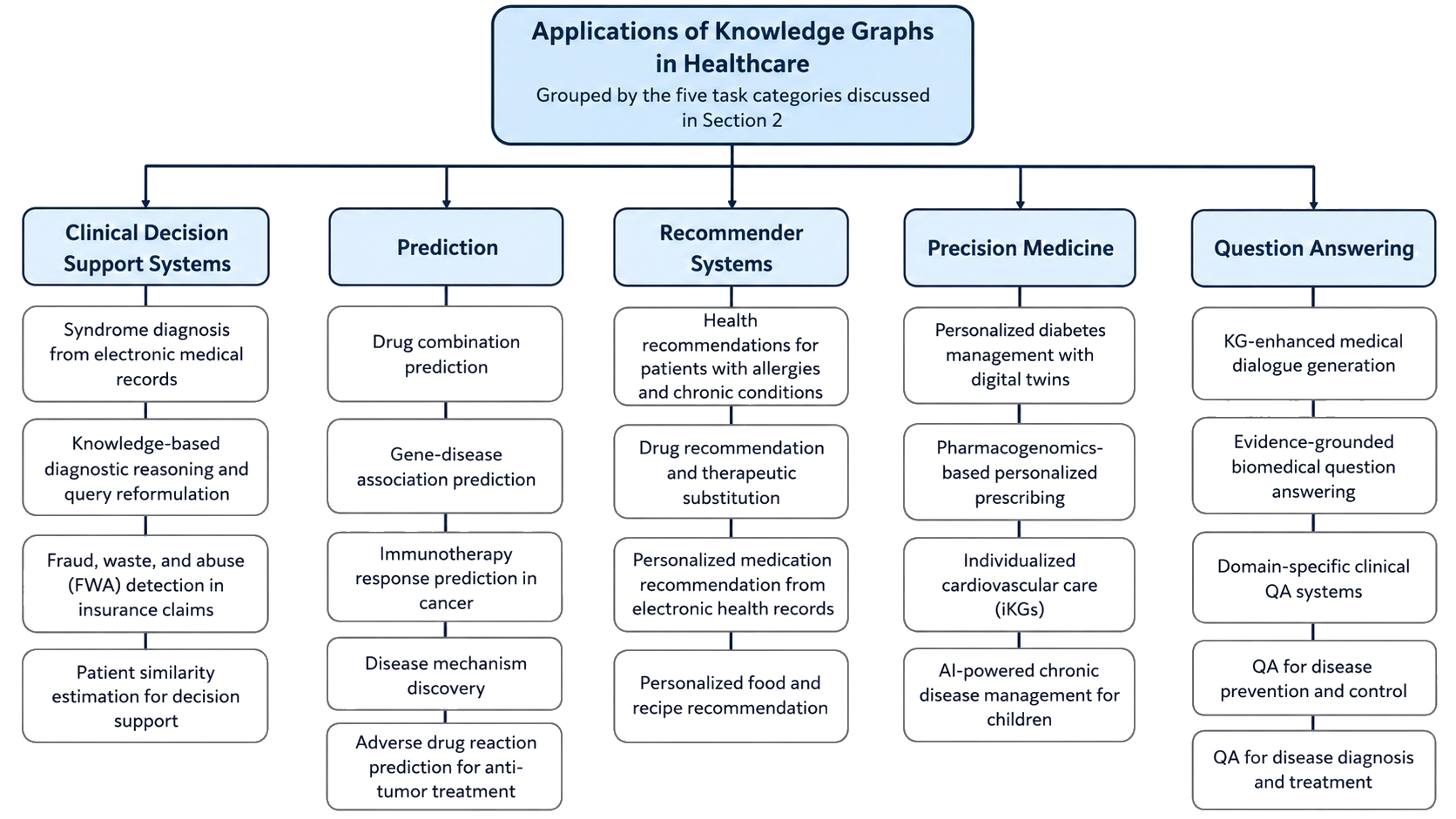}
\caption{Applications of knowledge graphs in healthcare, grouped into the five task categories discussed in Section 2. Each column lists representative types of KG-based tasks identified in the surveyed literature, such as syndrome diagnosis, drug combination prediction, personalized medication recommendation, digital-twin-based diabetes management, and evidence-grounded medical QA.}\label{fig:fig2}
\end{figure*}

\subsection{Clinical Decision Support Systems}
\label{subsec1}

Clinical Decision Support Systems (CDSS) aim to improve care quality, reduce medical errors, and lower healthcare costs by providing patient-specific recommendations based on existing medical knowledge. Traditionally, CDSS are divided into knowledge-based systems, which rely on manually encoded expert rules (e.g., if–then logic), and non-knowledge-based systems, which learn patterns directly from historical clinical data using artificial intelligence \cite{freidel2025knowledge}. Knowledge graphs (KGs) increasingly serve as a unifying representation for both paradigms: they encode curated medical knowledge and patient data in a semantically rich, machine-readable structure that supports logical reasoning, statistical learning, and hybrid approaches within CDSS.

A first group of KG-based CDSS focuses on diagnosis and the exploitation of clinical records. \cite{yang2022decision} develop a KG-based CDSS for syndrome diagnosis in Traditional Chinese Medicine (TCM). EMR entries are mapped to a TCM KG as incomplete triples, and embedding models such as ComplEx are used to infer missing entities and rank candidate syndromes, helping to disambiguate overlapping symptom patterns across syndromes. In a more general EHR setting, \cite{shang2021ehr} build a KG from heterogeneous clinical documents—patient records, laboratory results, physician notes—and combine natural language processing, machine learning, graph neural networks, and big data analytics to uncover latent relationships among patients, diseases, medications, and tests. Their system is positioned as a decision support tool that extracts underutilized information from routine clinical practice to enhance care quality and facilitate clinical research. Focusing on treatment recommendation, \cite{zhou2023corrigendum} introduce a CDSS for hypertension management based on a hyper-relational KG derived from Chinese treatment guidelines. The KG encodes diseases, symptoms, drug classes, drug types, and chemical compounds, and by combining advanced graph representation learning  with reasoning strategies over drug hierarchies, the system can complete missing elements in EMRs and recommend guideline-conformant drug regimens.

Several works emphasize KG-based reasoning and semantic query expansion as a way to support complex clinical decisions.  \cite{mohammadhassanzadeh2018investigating} propose SeDan, a semantics-based data analytics framework that combines RDF, RDFS, OWL, and SPARQL with probabilistic reasoning. When an initial query cannot be directly answered, SeDan reformulates it by exploiting hierarchical (generalization/specialization) and ordering relationships (e.g., interpolation, a fortiori reasoning), thereby generating alternative queries that can be answered by distributed resources such as MEDLINE, DrugBank, and Disease Ontology. This increases coverage while preserving explainability, as the reasoning patterns used to rewrite the query remain explicit. \cite{sun2021kgbref} introduce KGBReF, a KG-based framework for automatic, high-precision extraction of biomedical relations from large text corpora. By integrating sources such as UMLS, SNOMED CT, PubMed, DrugBank, and clinical guidelines into an extensive KG and defining “semantic trees” of high-level concepts, KGBReF uses NLP tools (e.g., XMedLan) and semantic reasoning to extract candidate relations—such as links between emotions and probiotics—which are then validated through expert annotation and inter-annotator agreement metrics. \cite{malik2020automated} propose a semi-automated pipeline that constructs a medical KG by extracting entities from research documents, mapping them to MeSH (or scispaCy annotations when MeSH is incomplete), organizing them into triples, and using BERT-based document embeddings with similarity metrics to predict links between entities. The resulting graph supports link prediction and document retrieval, enabling more focused evidence gathering for clinical questions and indirectly informing decision support.

KGs are also used to enrich patient representations and enable predictive CDSS. \cite{soman2023early} map EHR data from the UCSF Medical Center into the SPOKE KG and apply a modified PageRank algorithm (PSEV) to derive SPOKEsig embeddings—patient-specific signatures that capture both clinical and molecular context. A random forest classifier trained on these signatures predicts Parkinson’s disease (PD) onset one, three, and five years before the diagnostic code appears in the EHR, outperforming models based on raw EHR features. Importantly, SPOKEsig highlights interpretable prodromal patterns—such as sleep disturbances, behavioral changes, and autonomic dysfunctions—thus combining predictive performance with clinical explainability. \cite{lyu2023causal} construct a Causal Knowledge Graph (CKG) for diabetic nephropathy by extracting causal triples from SemMedDB, UpToDate, and Churchill’s Pocketbook of Differential Diagnosis and encoding them in OWL. EHR data are modeled in a hierarchical patient–visit–treatment structure and integrated into a Personalized Causal Knowledge Graph (PCKG). For each visit, depth-limited depth-first search retrieves causal paths (of length up to 3) to build a visit-level causal subgraph, providing individualized reasoning paths to support diagnostic decisions. \cite{lin2020patient} present DeepPS, a KG-driven patient similarity framework. They construct a medical KG from EHRs, ICD-9 codes, DrugBank, and physician-written descriptions (e.g., from Wikipedia), learn joint embeddings that align KG triples with textual embeddings (e.g., Skip-gram–based profiles), and represent each patient as a temporal sequence of medical entities. A Siamese CNN with spatial pyramid pooling computes patient similarity, improving treatment planning and hospital readmission prediction. At the discovery end of the spectrum, \cite{yang2022knowledge} introduce KGAP, a graph analytics platform that integrates LINCS omics signatures and IDG drug–gene–disease knowledge into a semantic KG. By applying shortest-path and centrality measures (closeness, betweenness, degree, eigenvector) on this graph, KGAP identifies genes associated with approved Parkinson’s drugs and highlights underexplored “Tdark” targets such as SYNGR3, which are then independently validated. While KGAP is primarily oriented toward target discovery and drug repurposing, it still represents a form of research-oriented decision support.

KG-based CDSS are not limited to physicians and clinicians; they can also support other stakeholders and system-level safety. \cite{sun2020medical} develop a CDSS for insurers that uses a large-scale medical KG—built from FDA drug labels, medical textbooks, and guidelines—to automatically identify suspicious cases of fraud, waste, and abuse (FWA) in insurance claims. Entities are extracted from unstructured texts using a deep-learning pipeline (BERT, BiLSTM, CRF with manual feature engineering and rule-based correction), and relations are learned with PCNNs and adversarial training. The unified KG then serves as a semantic backbone on which reasoning rules detect illogical prescriptions and diagnoses, successfully flagging around 70 percent of fraudulent cases and improving processing efficiency. 
More recent work explicitly combines KGs with large language models (LLMs) and hybrid AI to achieve interpretable and trustworthy decision support. \cite{vidal2025integrating} present a position paper on integrating medical KGs with neuro-symbolic and hybrid AI systems and introduce the TrustKG framework, which comprises layers for semantic data management, KG management, graph analytics, and human–machine communication. Within this architecture, VISE improves link prediction by combining SHACL constraints and mined Horn rules with KG embeddings, while HealthCareAI enriches KGs with semantic and causal information to support counterfactual reasoning about treatment outcomes in lung cancer. \cite{gao2025leveraging} propose DR.KNOWS, a diagnostic reasoning framework that augments LLMs with knowledge paths extracted from UMLS-based KGs. Clinical notes in SOAP format serve as input; subjective and assessment sections are used to retrieve relevant subgraphs via a Subgraph Isomorphism Graph Network and attention-based ranking, and the selected paths are injected into the LLM’s context to guide more informed and explainable diagnostic decisions. 

Overall, these studies illustrate that medical KGs can support CDSS along multiple dimensions: structuring EMRs and clinical narratives, encoding guideline knowledge, enabling semantic and causal reasoning, enriching patient representations for predictive modeling, and constraining or guiding LLM-based diagnostic agents. At the same time, important challenges remain. Many existing KGs are narrow in scope, focused on specific diseases or subdomains and built from limited sources, which restricts the generalizability of CDSS beyond their original context. Reasoning algorithms—whether symbolic, embedding-based, or hybrid—must be robust, scalable, and clinically validated to function as trustworthy assistants rather than opaque black boxes. There is also a need for broader coverage of general medical knowledge, better provenance tracking, and user-centered tools that make KG-driven recommendations transparent and actionable for different stakeholders \cite{chaoyu2020review}. Table 1 summarizes the main characteristics of the KG-based clinical decision support systems reviewed in this section, including their data sources, KG design, and target tasks.

\begin{table*}[!t]
\centering
\includegraphics[width=\textwidth]{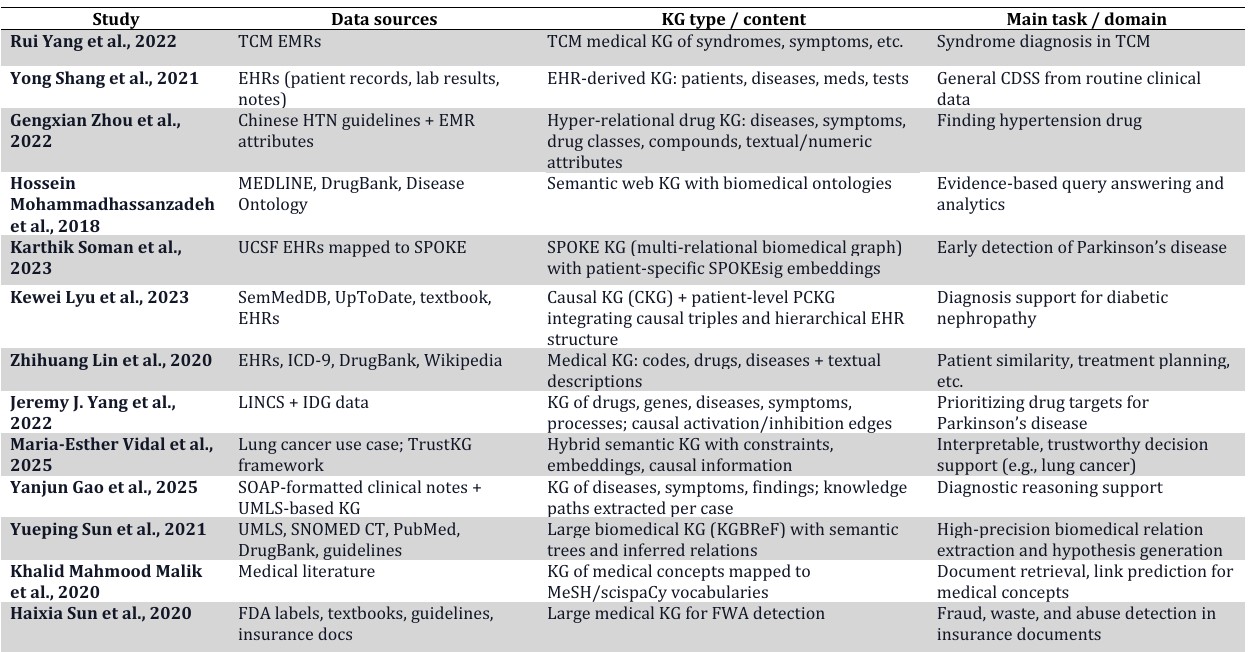}
\caption{Summary of representative knowledge graph–based clinical decision support systems . The table lists the main data sources, KG construction characteristics, and primary tasks or application domains for each study.}\label{tab:fig3}
\end{table*}

\subsection{Prediction}
One of the major applications of knowledge graphs in medicine is predictive modeling—such as forecasting disease onset, treatment response, adverse drug reactions, or gene–disease associations. A recurring challenge in these tasks is the lack of explainability in black-box models, which KGs can help mitigate by encoding semantic, relational, and sometimes causal structure over biomedical entities and patient data.

A first line of work exploits KGs to improve patient-level risk prediction from EHRs and clinical data. \cite{ma2018kame} propose KAME, a diagnosis prediction model that integrates EHR sequences with medical ontologies such as CCS and ICD. By learning embeddings for medical codes and their hierarchical ancestors and combining them with visit-level representations via a knowledge-guided attention mechanism, KAME captures both fine-grained and high-level disease concepts to predict future encounters more accurately. \cite{li2023prediction} construct a medical KG for predicting the risk of diabetic macular edema (DME) in diabetic patients. Thirty-nine risk factors, expanded into 116 qualitative intervals, are modeled as nodes with weighted edges derived from co-occurrence frequencies and association-enhancement algorithms; a generalized closeness metric is then used to compute individual risk scores on a 0 - 100 percent scale, providing an interpretable, KG-based risk assessment. To better understand polygenic diseases and optimize combination therapies, \cite{gao2023medical} introduce MK-GNN, a medical knowledge–guided GNN for drug combination prediction. Diagnostic histories and medication records from EHRs are used to derive heuristic medication features, which are processed through multi-head self-attention, while a drug interaction network is modeled using a GCN; patient features and graph-derived drug representations are then fused to predict effective drug combinations. \cite{tao2020mining} develop a heterogeneous health KG (HKG) from NHANES data, representing clinical test results, demographics, and medical histories as connected nodes and edges. Correlation analysis and clustering reveal latent health concepts, which are combined with domain-specific knowledge in a classification model that distinguishes healthy from unhealthy patients using a weighted fusion of the two knowledge types. \cite{xu2023seqcare} propose SeqCare, a framework for diagnostic prediction in EHRs that explicitly addresses label imbalance and noise in external KGs. Their three-stage pipeline—bias reduction via unsupervised graph contrastive learning, variance reduction through personalized KGs with adaptive graph adjustments, and a self-distillation phase that transfers knowledge from a teacher to a student model—produces robust representations and improves diagnostic prediction performance. \cite{wu2023megacare} introduce MEGACare, a multi-view hypergraph framework guided by medical domain knowledge. They enrich medical code representations with ontologies, code descriptions, and molecular data and model EHRs as hypergraphs in which each visit is a hyperedge over medical codes. By combining code graph, enhanced hypergraph, and sub-hypergraph views with optimized structures, MEGACare achieves strong results on both diagnosis prediction and medication recommendation.

A second line of research focuses on gene–disease and omics-driven prediction tasks. \cite{vilela2023biomedical} build a large KG from public resources such as Gene Ontology, DisGeNET, and Ensembl, encompassing more than 93,000 nodes and 1.7 million relations, and evaluate several KG embedding models (ComplEx, DistMult, TransE) for gene–disease association prediction. TransE with L2 norm achieves the best link ranking performance and uncovers biologically meaningful gene–disease subsets and signaling pathways, including mechanisms relevant to autism spectrum disorder, illustrating the utility of KGEs in variant prioritization and patient stratification. \cite{zhao2023biological} propose ICInet, a GNN-based model for immunotherapy response prediction that integrates prior biological knowledge from gene regulatory and molecular interaction networks into a graph-learning framework driven by gene expression profiles. An attention module highlights critical network regions, such as paths near therapeutic targets, yielding both improved predictive performance and interpretable biomarkers of treatment response. \cite{renaux2023knowledge} develop BOCK, a KG that unifies biological networks, ontologies, and data on oligogenic diseases to investigate oligogenic disease mechanisms. Their ARBOCK method uses meta-paths as features and association rule learning to predict causal gene–gene interactions, presenting results as interpretable subgraphs to facilitate expert-driven hypothesis generation. \cite{choi2021identifying} introduce KGED, a GNN-based model that jointly embeds a biomedical KG—built from CTD, BioGRID, and MalaCards with relations such as chemical–gene, chemical–disease, gene–gene, disease–gene, and disease–symptom—and textual descriptions of entities. A convolution-based scoring function is used to infer gene–disease links, and KGED outperforms TransE and ConvKB in link prediction while identifying central disease-related genes, particularly in oncology. \cite{lan2022kgancda} propose KGANCDA, a GAT-based model over a multi-entity biomedical KG (circRNA, disease, miRNA, lncRNA) to predict circRNA–disease associations. By aggregating both direct and high-order neighborhood information across multi-source data and feeding learned representations into an MLP, KGANCDA addresses sparsity and limited low-order interactions, achieving improved prediction of circRNA–disease links.

A third set of studies applies KGs to drug discovery, drug repurposing, and pandemic-related mechanistic analysis. As summarized by \cite{zeng2022toward}, KGs support tasks ranging from molecular property prediction and structure optimization to drug repurposing and adverse drug reaction (ADR) prediction. \cite{ma2023kgml} present KGML-xDTD, a KG-based machine learning framework that combines a drug repurposing predictor (DRP) with a mechanism-of-action (MOA) explainer. GraphSAGE is used to embed nodes in a biomedical KG enriched with PubMedBERT-derived textual features, and a random forest classifier labels drug–disease pairs as treatable or untreatable. For explanation, an actor–critic reinforcement learning module with demonstration-guided path sampling extracts biologically plausible reasoning paths within the KG, linking predictions to mechanistic evidence. \cite{che2021knowledge} construct a COVID-19-centered biomedical KG by adding viral targets (e.g., RdRp, ACE2, pp1ab, pol) extracted from recent literature and propose Att-GCN-DDI, a GNN with attention that learns node features and reconstructs the drug–disease interaction matrix to identify drug–disease interactions. Their model not only achieves high accuracy on standard DDI tasks but also recovers several clinically validated COVID-19 drugs. \cite{zhu2020knowledge} build a comprehensive drug KG from PharmGKB, TTD, KEGG DRUG, DrugBank, SIDER, and DID, using meta-path–based representations and KG embeddings as features for ML models that predict candidate drugs for diseases such as diabetes mellitus. The use of meta-paths enhances both accuracy and interpretability by exposing recurrent patterns in drug–disease relationships. \cite{kramer2021coronavirus} develop the Coronavirus Network Explorer (CNE), a large-scale causal KG that connects SARS-CoV-2 viral proteins to host proteins, biological processes, diseases, and pathways through activation/inhibition relations. Embedding-based models that account for interaction directionality identify key pathways—such as blood coagulation, inflammation, and IL-6 signaling—and candidate drugs whose overlap with clinical trial compounds supports the KG’s relevance for drug repurposing.

A final, extensive line of work uses KGs to model drug–target interactions (DTIs), drug–drug interactions (DDIs), drug similarity, and ADRs. \cite{mohamed2020discovering} introduce TriModel, a tensor factorization–based approach that formulates DTI prediction as link prediction on a KG built from DrugBank, KEGG, and related sources. By learning embeddings for drugs and proteins and computing interaction scores, TriModel captures both on-target and off-target effects. \cite{zhang2021discovering} propose MHRW2Vec-TBAN for predicting DTIs and DDIs on KGs constructed from DrugBank, KEGG, and PharmGKB. MHRW2Vec uses Metropolis–Hastings random walks combined with Word2Vec to generate structure-aware node embeddings, which are then fed into TBAN, a deep architecture that integrates TextCNN, Bi-LSTM, and attention to model local features, long-range dependencies, and salient patterns for interaction prediction. \cite{shen2019kgdds} design KGDDS, a system for drug similarity estimation and therapeutic substitution. They construct a drug KG from DrugBank, SIDER, NDF-RT, and literature, represent drugs via chemical fingerprints and DRKG-based ComplEx embeddings, and combine multiple feature channels—including side effects, mechanisms, and textual descriptions—through attention mechanisms. KGDDS outperforms classical similarity models (e.g., GADES, Res, Wpath, Hybrids, MedSim, Tiresias) in Pearson and Spearman correlation and offers visualization tools for interactive exploration. \cite{hao2023enhancing} address ambiguity in DDIs by combining chemical structure features and DRKG embeddings within a three-way decision framework that partitions predictions into interacting, non-interacting, and uncertain boundary regions. An enhancing module based on a graph-enhanced convolutional neural network refines decisions in the boundary region, enabling nuanced uncertainty handling and stronger DDI prediction performance. \cite{su2022attention} propose DDKG, an attention-based GNN framework in which initial drug representations from attribute encoders (e.g., SMILES-based) are iteratively updated by propagating information over a drug KG. Interaction probabilities are then computed via the inner product of learned embeddings, enabling end-to-end DDI prediction. \cite{ye2021unified} introduce KGE-NFM, which first applies KG embedding models such as DistMult to a multi-entity biomedical KG (drugs, proteins, diseases, side effects), then uses PCA to denoise and reduce embedding dimensionality, and finally feeds the refined embeddings into a Neural Factorization Machine to capture nonlinear feature interactions for DTI prediction. \cite{wang2021adverse} construct a Tumor–Biomarker KG (TBKG) from MEDLINE, linking tumors, biomarkers, drugs, and ADRs via weighted relations. Naive Bayes is used to assess tumor–biomarker relationships, and PU-learning–based classifiers (SVMs, decision trees, random forests) predict ADRs for anti-tumor therapies; explanatory paths such as tumor–biomarker–drug provide mechanistic insight into adverse events. \cite{zhang2021prediction} also target ADR prediction by building a KG from DrugBank and SIDER that includes drugs, side effects, drug targets, and clinical symptoms. Using Word2Vec (CBOW) to embed entities and training a logistic regression model on difference vectors between drug and ADR embeddings, they achieve accurate classification of causal drug–side-effect links. Finally, \cite{munoz2019facilitating} adopt a multi-label learning perspective on ADR prediction, integrating SIDER, DrugBank, PubChem, and KEGG into a KG and evaluating multiple classifiers (decision trees, random forests, kNN, MLPs, logistic regression) to rank ADRs per drug, explicitly modeling the inherently multi-label nature of drug safety profiles.
Overall, these studies demonstrate that predictive models grounded in healthcare KGs can capture rich semantic and relational structure that is difficult to learn from flat features alone. By leveraging ontology hierarchies, molecular interaction networks, multi-modal biomedical data, and patient-level graphs, KG-based approaches achieve competitive or superior performance across diverse tasks—from disease risk prediction and treatment response to gene–disease association, drug repurposing, DTI/DDI modeling, and ADR forecasting. Importantly, many of these methods provide interpretable outputs, such as risk scores tied to specific factors, attention-highlighted network regions, meta-path explanations, or explicit graph paths that connect drugs, targets, and phenotypes. Nevertheless, challenges remain regarding KG completeness, bias in underlying data sources, robustness of graph learning methods, and the need for systematic clinical validation before these predictive models can be routinely trusted in high-stakes medical decision-making. Table 2 summarizes the KG-based prediction models reviewed in this section, including their data sources, graph representations, and target clinical or biomedical tasks.

\begin{table*}[!t]
\centering
\includegraphics[width=\textwidth]{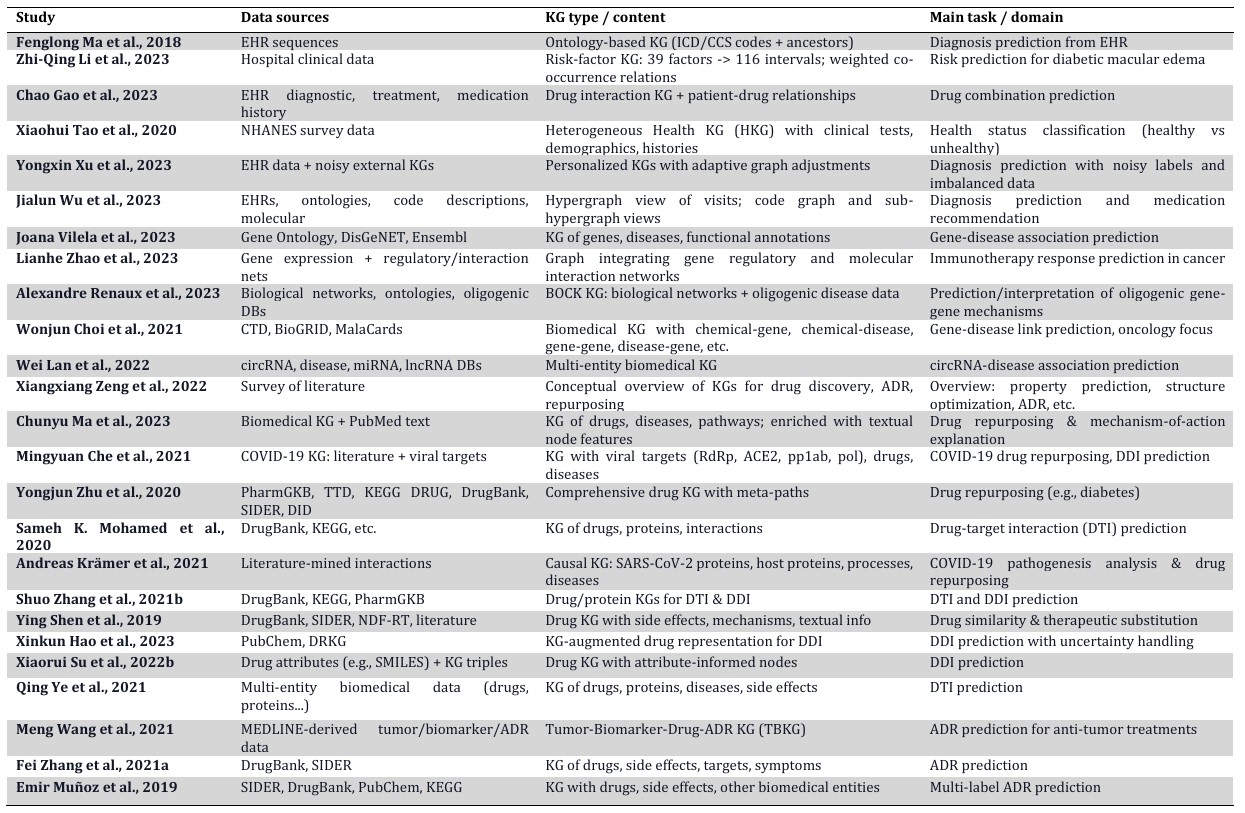}
\caption{Summary of representative knowledge graph–based prediction models in medicine.}\label{tab:fig4} 
\end{table*}
\subsection{Recommender Systems}
Health recommender systems (HRS) typically target two main user groups: (i) patients or healthy individuals seeking personalized suggestions on medications, nutrition, physical activity, and lifestyle choices; and (ii) healthcare professionals who rely on such systems for treatment options, diagnostic support, and clinical decision assistance. In both cases, explainability is crucial: users must be able to understand and trust the rationale behind recommendations, especially when they affect health-related behavior or clinical care \cite{tran2021recommender}. Knowledge graphs provide a natural backbone for explainable recommendation by explicitly linking users, symptoms, diseases, treatments, and lifestyle factors through semantic relations and enabling path-based justifications.

Several works use KGs as integrative layers for heterogeneous health data and rule-based recommendation. \cite{gyrard2022reasoning} develop a personalized health KG framework that integrates sensor data, clinical records, environmental conditions, and other sources to support tailored recommendations for patients with allergies and chronic conditions. Using semantic web technologies (RDF, OWL, Apache Jena), they annotate raw data, transform them into structured triples, and apply rule-based inference to derive high-level concepts such as “high pollen exposure.” SPARQL queries over the inferred KG then power personalized alerts and recommendations. At a more generic level, \cite{wang2019knowledge} highlight that classical collaborative filtering (CF) methods suffer from sparsity and cold-start issues in health and recommendation domains; they advocate the use of personalized KGs to incorporate richer user-specific and item-specific signals, thereby alleviating these problems and providing a semantic basis for recommendations.

A second group of studies focuses on medication and treatment recommendation for patients. \cite{huang2024interpretable} propose an explainable graph-based drug recommendation framework for chronic disease management, particularly in elderly patients. Their system integrates medication histories, demographic information, test results, and drug usage logs, and explores two graph collaborative filtering variants: GCF-YA, which employs graph attention to weight neighbors and emphasize informative user–drug interactions, and GCF-NA, which uses standard information propagation for settings where simpler neighborhood modeling is sufficient. By operating over a graph that encodes relationships among patients, drugs, and clinical factors, the framework can exploit higher-order connections while exposing attention weights as an explanation of why specific drugs are recommended. \cite{hamane2023healthpathfinder} introduce HealthPathFinder, a personalized recommendation framework that combines transactional data (e.g., patient–drug interactions) with semantic knowledge from a medical KG. The system extracts positive and negative multi-hop paths between patients and candidate treatments and employs multi-head attention to rank these paths, enabling both accurate predictions and path-based explanations (e.g., patient → diagnosis → drug) that justify recommended therapies. \cite{elahi2024knowledge} present KGCAN, a responsible, KG-based recommendation model that treats the KG as auxiliary information to enrich item representations. Through heterogeneous knowledge propagation, multiple graph neural layers, and mechanisms such as a Contextualized Attention-Aware Embedding and a User-Specific Attention module, KGCAN aggregates signals from first- and higher-order neighbors and dynamically adapts its focus to each user’s historical behavior and preferences, supporting personalized and potentially more “responsible” recommendations. \cite{bhoi2021personalizing} propose PREMIER, a personalized medication recommender that leverages EHRs and graph-based models. PREMIER constructs a patient query vector from diagnoses, procedures, and prior medications using attention-based neural encoders, and then queries two separate GAT-modeled graphs: one capturing medication co-occurrence and another representing drug–drug interactions (DDIs). In addition to recommending medications, PREMIER quantifies the contributions of diagnoses, procedures, previous drugs, and graph-level information, thereby enhancing the transparency of its outputs. In the Traditional Chinese Medicine (TCM) domain, \cite{wang2020novel} builds a TCM KG from web sources containing diseases, prescriptions, herbs, and symptoms and uses Node2vec to embed the graph into a vector space. A k-nearest neighbors approach in this embedding space is then applied to recommend TCM prescriptions based on patient symptom profiles, offering a simple yet interpretable KG-based prescription retrieval mechanism.

A third line of work targets lifestyle and nutrition recommendation, where KGs encode relationships among foods, nutrients, diseases, and user preferences. \cite{xu2024developing} develop ElCombo, a personalized food recommender for older adults in Chinese communities who face malnutrition risk, chronic illnesses, and low nutritional literacy. They construct a FoodKG from recipe websites, communal dining choices, and food composition tables, and build user profiles from surveys and IoT data. Recommendation algorithms then account for both health constraints (e.g., disease-specific dietary restrictions) and personal taste, illustrating how KGs can coordinate clinical and preference-based requirements in food recommendation. \cite{huang2022novel} design a personalized recipe recommender that leverages a nutritional KG and CF techniques. By integrating web-scraped recipes, ingredient information, and disease-related dietary guidelines, their system generates health-aware, customized recipe suggestions, while addressing cold-start and data fragmentation issues through the structured representation of food and nutrition knowledge. \cite{li2023health} propose a dual-KG approach that explicitly separates user taste and health considerations. One KG captures preference patterns based on user behavior, food names, and ingredients, while a second KG encodes health-related information based on nutritional scores and standards such as FSA. Graph convolutional networks are applied to the preference graph and graph attention networks to the health graph; a knowledge transfer mechanism and a back-transfer strategy align and synchronize the two embedding spaces, enabling the system to recommend options that balance palatability with nutritional quality. \cite{chi2018knowledge} address multi-modal knowledge management for sustainable healthcare by introducing a TCM-inspired dietary guidance system. They build a KG with five main concept types—foods, meals, nutrients, disease symptoms, and target groups—and relations such as good-for, bad-for, related, and different, using data from trusted health and nutrition platforms. Machine learning models (e.g., CRF, SVM, BiLSTM, Naive Bayes, KNN) are used for entity and relation extraction, and the resulting KG powers multi-dimensional semantic search and personalized dietary recommendations through both natural language queries and image-based interfaces.

Finally, several KG-based frameworks for recommendation emphasize general modeling patterns rather than a specific application domain. In many of these systems, graph neural networks and attention mechanisms play a central role in propagating information across KGs, weighting relevant neighbors, and capturing high-order interactions while preserving a path-based structure that can be surfaced as an explanation. Collaborative filtering is often reformulated as a recommendation over a user–item–knowledge graph, where KGs alleviate sparsity and cold-start by linking users and items to a rich network of auxiliary entities (e.g., diseases, symptoms, drugs, foods, nutrients) and enabling semantic generalization beyond direct interaction histories.

Overall, the reviewed studies show that KG-based health recommender systems can provide both accurate and highly explainable recommendations across a spectrum of use cases—from medication selection and TCM prescriptions to diet planning and lifestyle guidance. By leveraging structured relationships among clinical concepts, user attributes, and multi-modal data sources, these systems can move beyond shallow co-occurrence patterns and offer recommendations grounded in explicit medical or nutritional knowledge. At the same time, important challenges remain, including handling incomplete or noisy KGs, aligning clinical guidelines with individual preferences, ensuring fairness and responsibility in personalized recommendations, and evaluating long-term behavioral and clinical impact of KG-driven recommendations in real-world settings.

\subsection{Precision Medicine}
Precision medicine, or personalized medicine, aims to tailor prevention, monitoring, and treatment strategies to individual characteristics such as lifestyle, comorbidities, demographics, and molecular profiles. Knowledge graphs are particularly well suited to this paradigm because they can integrate heterogeneous data sources—EHRs, omics, guidelines, sensor data, and patient-reported outcomes—into a coherent, semantically rich representation that supports both personalization and explainability.

Several works explicitly construct personalized or individual-level KGs to support precision care. \cite{sarani2024personalized} propose a framework for personalized diabetes management that combines digital twin technology with a Personalized Health Knowledge Graph (PHKG). Data from medical history, clinical measurements, lifestyle, and dietary preferences are harmonized using HL7 standards and an explicit ontology, and then linked in a PHKG for each patient. A digital twin instantiated from this PHKG feeds simulation models and machine learning algorithms that predict blood glucose trajectories, optimize insulin dosage, and recommend lifestyle and diet adjustments, thereby improving glycemic control and patient engagement over time. Figure 3 provides a conceptual illustration of how a patient-centric medical knowledge graph can be instantiated as a digital twin, integrating diagnoses, laboratory results, medications, lifestyle factors, biomarkers, and demographic attributes into a unified representation for simulation and prediction. \cite{ping2017individualized} introduce the concept of individual Knowledge Graphs (iKGs) for cardiovascular medicine. Their vision is to fuse biological knowledge, patient histories, and outcomes into a unified patient-level graph that supports both translational research and clinical practice. The proposed dual-branch architecture—comprising a Translational Research Arm and a Clinical Practice Arm—positions iKGs as a bridge between discovery and bedside decision-making, and as a substrate for future intelligent virtual assistants that help clinicians implement personalized therapies.

Other approaches focus on personalized prescribing and interpretable disease trajectory modeling. \cite{jamrat2023precision} develop a knowledge-driven clinical decision support tool (CDST) for personalized prescribing that integrates genetic and non-genetic factors, including drug–drug and herb–drug interactions, lifestyle habits, and liver/kidney function. Their KM2PS model aggregates pharmacogenomic databases, clinical guidelines, and pharmaceutical resources into a semantic network, then uses automated screening logic to generate Personalized Prescribing Reports based on patient genotypes and clinical parameters, enabling clinicians and pharmacists to weigh efficacy and safety at the individual level. \cite{yang2023kerprint} present kerprint, a framework that combines deep learning with KGs to enhance both performance and interpretability in disease diagnosis prediction. For each patient, a sequence of visits with diagnoses, medications, labs, and procedures is extracted from EHRs; a local medical KG is built per visit using breadth-first search, capturing quadruples (head entity, relation, tail entity, timestamp) to preserve temporal structure. An element-wise attention mechanism aligns relevant subgraphs from a global medical KG with the patient’s local graph, supporting retrospective interpretation (highlighting historical pathways that contributed to disease progression) and prospective interpretation (forecasting likely future trajectories when a patient enters a new disease phase). These capabilities allow clinicians to track and explain changes in disease severity over time. \cite{yu2022improving} design AICMS, an AI-powered chronic disease management system for children that integrates KGs, IoT devices, and big data analytics. Wearable sensors collect physiological and activity data via a patient client, while a hospital client provides clinicians with analytic reports; a central processing unit hosts AI modules for follow-up, chatbot interaction, and diet recommendation. A medical KG links diseases, symptoms, and encounters, enabling the system to detect critical health patterns, trigger alerts in urgent cases, and generate context-aware recommendations for long-term disease management.

Taken together, these studies demonstrate how KGs can act as the structural core of precision medicine systems by: (i) organizing heterogeneous clinical, behavioral, and molecular data into personalized graphs (PHKGs, iKGs); (ii) encoding pharmacogenomic and clinical guideline knowledge for individualized prescribing; and (iii) providing a substrate on which deep learning and GNN-based models can perform temporally aware and knowledge-informed prediction. At the same time, important challenges remain, including maintaining up-to-date and consistent knowledge across rapidly evolving biomedical domains, ensuring interoperability across institutions and devices, quantifying the clinical impact of KG-driven digital twins and personalized graphs, and responsibly integrating LLMs and multimodal data (e.g., imaging, wearable streams) into precision workflows without compromising safety, privacy, or equity.

\begin{figure*}[t]
\centering
\includegraphics[width=.9\textwidth]{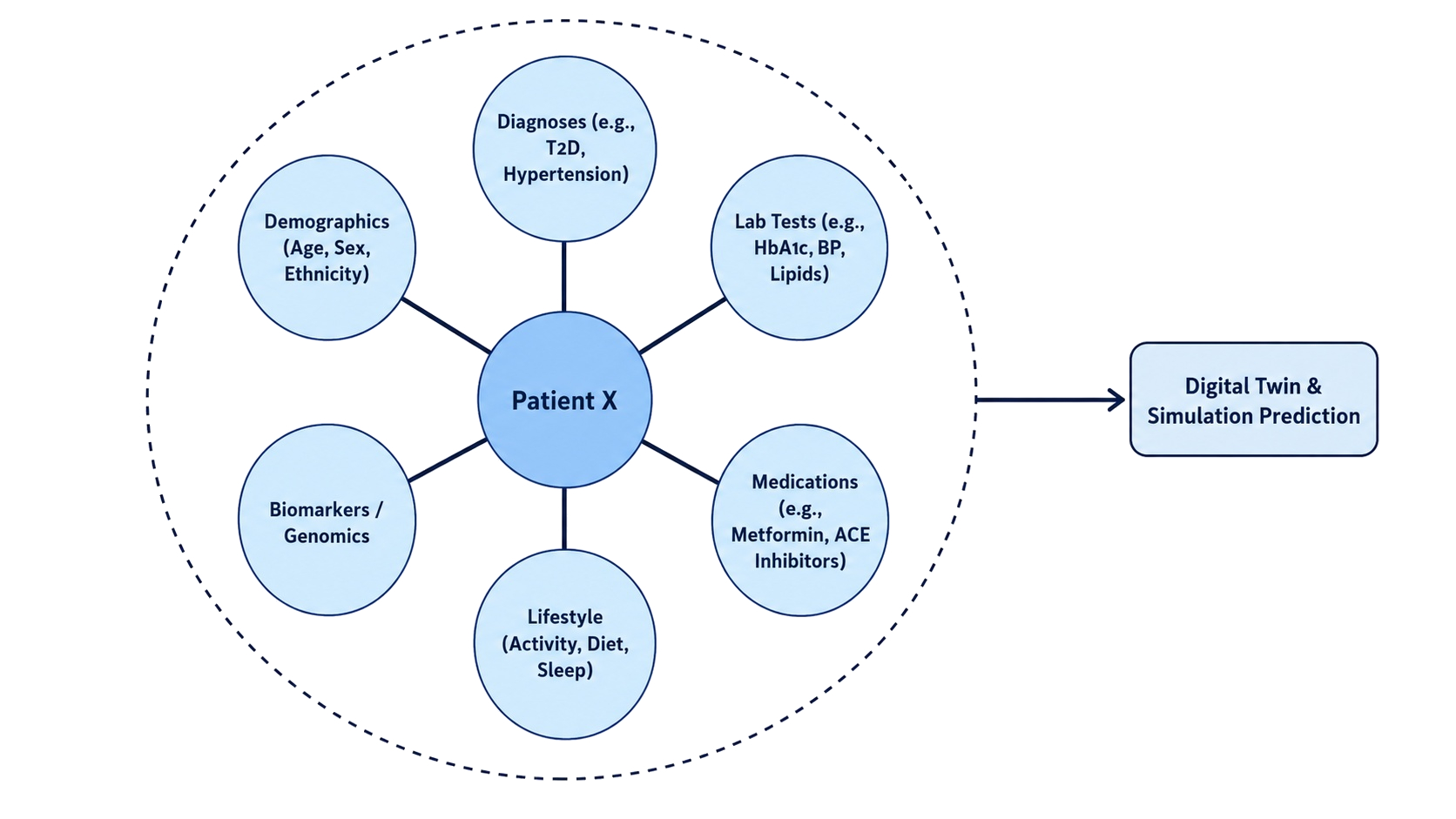}
\caption{Conceptual view of a patient-centric medical knowledge graph and its use as a digital twin. Clinical concepts such as diagnoses, laboratory tests, medications, lifestyle factors, biomarkers/genomics, and demographic attributes are linked around an individual patient node, and the resulting graph is leveraged for simulation and prediction of future health trajectories.}\label{fig:fig5}
\end{figure*}
\subsection{Question Answering (QA)}
With the rapid advancement of natural language processing (NLP) and the emergence of large language models (LLMs), medical question answering (QA) has become an increasingly important interface for accessing healthcare knowledge. In this high-stakes domain, accuracy, grounding in reliable sources, and explainability are essential. Knowledge graphs provide a structured, semantically rich substrate that can be combined with NLP models and LLMs to support fact-checked, interpretable answers rather than uncontrolled free-text generation.
Several works explicitly combine KGs with neural language models for medically grounded dialogue and text generation. \cite{varshney2023knowledge} present MedKgConv, an end-to-end medical dialogue generation model that integrates UMLS-based KGs into a transformer encoder–decoder architecture. Concepts and relations for diseases, symptoms, and lab tests are extracted as (head, relation, tail) triples, and graph-derived features are injected via a multi-head attention mechanism (MedFact) and a gated copy module, enabling the model to generate clinically relevant responses that are aligned with KG content rather than relying solely on language modeling priors. \cite{soman2024biomedical} introduce KG-RAG, a retrieval-augmented generation framework that links disease-related entities in user queries to nodes in the SPOKE KG. Relevant subgraphs are pruned and verbalized into concise, prompt-aware contexts that are appended to the input of LLMs such as Llama-2-13b, GPT-3.5-Turbo, and GPT-4. KG-RAG improves answer accuracy and robustness to small input perturbations while reducing token usage compared to Cypher-based querying or full-text indexing, illustrating how domain KGs can help control context length and keep generation grounded in biomedical evidence.

Another line of work builds KG-centric QA systems on top of EMRs or general medical KGs, often with explicit attention to interpretability and clinician oversight. \cite{sheng2020dsqa} propose DSQA, a domain-specific QA system that uses EMRs as its primary data source and incorporates a “doctor-in-the-loop” design. A conceptual KG encoding diseases, drugs, and symptoms is instantiated with patient-level data, and clinicians help define causal relations that form an explicit causal graph. DSQA uses this structure both to generate QA pairs for training and to answer queries at inference time, combining traditional ML models (e.g., SVM, Seq2Seq) with graph reasoning and visual subgraph explanations so that users can inspect the causal paths underlying each answer. \cite{zhang2021knowledge} builds a Chinese medical QA system based on a KG containing diseases, drugs, symptoms, departments, and tests, using data from open sources and the Xunyiwenyao platform. User queries are segmented and represented with TF-IDF, classified via Naive Bayes, and then matched to KG entities; answers are retrieved using Neo4j and presented through an interactive Chinese interface. \cite{shuai2019research} follow a similar KG-based design: they construct a Neo4j medical KG (diseases, drugs, symptoms, departments), use TextCNN for question-type classification, perform jieba-based segmentation with a custom dictionary for keyword extraction, and execute Cypher queries to return precise answers. These systems highlight a common pattern: KGs provide a schema and instance layer for medical entities, while relatively lightweight NLP models handle question understanding and mapping onto graph queries.

Disease and domain-specific QA systems further demonstrate the flexibility of KG-based approaches. \cite{sun2021intelligent} develop a myopia-focused QA system for prevention and control in children and adolescents. Data from Baidu Encyclopedia, Chinese Wikipedia, and specialized medical websites are processed via semi-structured extraction and pattern-based entity recognition to produce a Neo4j KG with entities such as myopia, symptoms, treatments, and preventive methods. A pipeline consisting of BERT-based question understanding, Cypher queries, and answer generation supports both text and voice interaction, delivering targeted answers to questions like “What are the symptoms of myopia?” or “How can myopia be prevented?”. \cite{yin2022question} design a KG-based QA system for hepatitis B diagnosis and treatment in TCM. Structured information from EMRs and web pages is stored in a TCM KG; Bi-LSTM+CRF is used for named entity recognition, Word2Vec-based pattern matching interprets queries, and Cypher queries retrieve the relevant knowledge, producing precise answers grounded in both modern and traditional medical sources. In the mental health domain, \cite{gunjan2022retracted} propose a QA platform for psychological counseling and suicide risk assessment. Data from social media and medical forums are processed using NLP methods, features are selected via statistical tests such as Chi-square, and multiple ML models are evaluated, with a Bi-LSTM chosen for its contextual understanding of sequential text. Cloud-based KGs serve as storage and retrieval layers, providing structured knowledge that can aid mental health professionals in quickly identifying high-risk behaviors and supporting decision-making.

Recent work in TCM demonstrates how KGs and LLMs can be tightly coupled for both QA and generative tasks. \cite{zhuang2025tcm} introduce TCM-KLLaMA, an intelligent prescription generation model that integrates a TCM KG with a Chinese LLaMA2-based LLM. The KG encodes relationships among symptoms, tongue and pulse diagnoses, and herbs, while a Synonym and Matching Knowledge Injection (SMKI) mechanism uses synonym lexicons and sentence-embedding-based fuzzy matching to map free-text symptom descriptions to KG entities. Instead of unconstrained text generation, the model replaces the standard LM head with a linear layer and sigmoid output over 811 herbs, turning prescription generation into a multi-label classification problem that restricts outputs to valid ingredients. LoRA-based fine-tuning on 33,765 symptom–prescription pairs yields performance gains over topic-model and KG-enhanced baselines, and ablation studies confirm the contributions of both SMKI and the constrained output layer to improved accuracy and reduced hallucination. \cite{duan2025research} develop a TCM case-based QA system that integrates LLMs with a KG of historical medical cases from Medical Cases of Wang Zhongqi. A four-stage pipeline—schema design based on national TCM standards, LLM-based NER and relation extraction (GLM-3-Turbo), LLM-assisted normalization with expert proofreading, and Neo4j-based storage and visualization—produces a structured case KG with five key entity types (disease, symptom, pathogenesis, treatment principle, medication). On top of this KG, GPT-3.5-Turbo (orchestrated via LangChain) parses user questions, fills Cypher templates, queries the graph, and generates natural-language answers grounded in retrieved subgraphs. Evaluation with RAGAS metrics (faithfulness, answer relevance, context recall) and expert SUS-style assessments shows that the KG-augmented system outperforms standalone LLMs in safety, usability, and fluency, underscoring the value of structured TCM knowledge for trustworthy clinical QA.

KG-based QA paradigms are also applied to nutrition and food recommendation, blurring the line between QA and recommendation. \cite{zeng2020exploiting} propose a conversational system that elicits user food preferences through dialogue. A large food KG built from Freebase is combined with topic embeddings derived from Wikipedia to capture topic correlations; when Freebase lacks specific entities, completion methods such as TransE are used to fill gaps. The system actively asks follow-up questions to refine its understanding of a user’s dietary preferences, using the KG as a scaffold for multi-turn preference elicitation. \cite{chen2021personalized} introduce pFoodReQ, a personalized food recommendation framework that formulates health-aware food search as a constrained QA problem over a FoodKG. Natural-language queries enriched with user constraints (e.g., allergies, nutritional needs, consumption history) are processed with query expansion and KG augmentation; entities are tagged as positive or negative constraints during embedding and query generation. Subgraph extraction over the FoodKG returns recipes that satisfy both preference and health constraints, and evaluation indicates substantial gains over non-personalized baselines.

Overall, the reviewed studies show that combining KGs with QA and dialogue systems enables a spectrum of health applications—from general medical question answering, disease-specific QA, and TCM consultation to psychological counseling support and food/nutrition assistance. Across these systems, KGs play several recurring roles: serving as a schema and instance layer for medical entities, providing explicit causal or semantic relations that can be visualized as subgraphs, constraining or enriching LLM-based generation, and supporting path-based explanations of answers. Despite these advances, important challenges remain, including maintaining up-to-date and trustworthy KGs, handling ambiguous or noisy user queries, integrating multimodal data such as imaging or physiological signals, and establishing robust evaluation frameworks that capture not only answer accuracy but also safety, bias, and real-world clinical impact.

\section{Knowledge Graph Generation}
\label{sec:sec3}

Given the wide range of applications for medical knowledge graphs (KGs) in healthcare and biomedical research, an essential question is how to construct high-quality, domain-specific KGs that are robust, extensible, and suitable for downstream tasks. In practice, KG generation frameworks must address several recurring challenges: integrating heterogeneous data sources, performing reliable entity and relation extraction, aligning concepts across terminologies, and enriching or pruning the resulting graph to balance coverage, accuracy, and interpretability. This section reviews representative frameworks and highlights their design goals and methodological choices. Figure 4 summarizes a generic pipeline for constructing medical knowledge graphs from heterogeneous data sources, which provides a common conceptual backbone for the frameworks reviewed in this section.\cite{zhang2020hkgb} introduce HKGB, a comprehensive and flexible framework for constructing healthcare KGs from diverse and largely unstructured sources such as EHRs, health standards, and existing databases. HKGB combines machine learning with a clinician-in-the-loop strategy to ensure semantic correctness and clinical relevance. The architecture includes tools such as HL-inspection for detecting new concepts, HL-annotation for annotating unstructured texts, HL-synonym for normalizing variant medical terms, and HL-rule for defining rule-based information extraction patterns. Beyond transforming raw data into triples, HKGB explicitly tackles operational challenges like data synchronization and concept integration, supporting scalable and extensible KG construction in real clinical environments. \cite{chen2019robustly} focus on building a health KG from more than 270,000 emergency visits to learn causal relationships between diseases and symptoms. Their work emphasizes robustness and generalizability: they systematically evaluate the accuracy of inferred relations, identify error sources such as small sample sizes and hidden confounders, and use nonlinear functions and the do-operator for causal inference and hypothesis testing. Taken together, these frameworks demonstrate how careful pipeline design and attention to causality can turn EHR data into clinically meaningful, robust KGs.

Deep learning–based information extraction has motivated end-to-end KG construction pipelines that automate large portions of the process. \cite{li2022deepkg} present DeepKG, a comprehensive framework for automatically extracting biomedical knowledge from text and representing it as triples. At its core lies CHIEF (Cascaded Hybrid Information Extraction Framework), which combines multiple deep models to perform precise biomedical triple extraction (entity–relation–entity). AutoML-based optimization modules such as AutoTransX are then used to refine relation representations and improve link prediction performance. DeepKG is designed not only to build KGs but also to support downstream applications—including drug discovery, clinical decision support, and QA—illustrating how tightly coupled extraction and reasoning components can form a general-purpose biomedical KG platform.

A number of frameworks explicitly target the transformation of clinical narratives into structured KGs for decision support. \cite{thukral2023knowledge} propose a pipeline that converts free-text clinical narratives into KGs tailored for clinical decision support systems. The process includes sentence decomposition (e.g., using NLTK and POS tagging), NER-based entity recognition, and classification against multiple biomedical ontologies such as Symptom Ontology, Disease Ontology, RadLex, FMA, and demographic-specific ontologies. Extracted entities and relations are then structured as triples and integrated into a clinical KG. Importantly, the framework supports ontology enrichment: novel concepts identified in clinical text can be proposed and validated by domain experts before being incorporated into the ontological backbone, ensuring that the KG remains both robust and evolvable. \cite{alam2023automated} design an automated clinical KG generation framework to support evidence-based medical decision-making. Their system combines concept extraction, semantic enrichment, optimized clustering, and RNN-based deep learning models. Using BioBERT embeddings, it identifies PICO elements (Participants, Interventions, Comparisons, Outcomes) and extracts relationships among them, automatically generating topic-focused subgraphs (e.g., for COVID-19 or cerebral aneurysms) that can be searched and explored efficiently. These subgraphs function as condensed, evidence-centric KGs that streamline retrieval in rapidly evolving clinical domains.
Semantic web technologies and ontology-centric designs also play a central role in KG generation. \cite{shi2017semantic} address the challenge of organizing and integrating textual medical knowledge (TMK) from heterogeneous and rapidly growing sources. They propose the Semantic Health Knowledge Graph (SHKG), which uses RDF and OWL standards to unify TMK with hospital information system (HIS) records. A contextual inference pruning algorithm is introduced to remove spurious or meaningless inferences, thereby improving precision and recall of reasoning tasks over the KG. SHKG exemplifies how semantic web standards, when combined with targeted pruning, can produce KGs that are interoperable yet computationally manageable for inference.

Finally, several frameworks focus on lifestyle- and nutrition-related KGs that, while domain-specific, illustrate general design principles for health KG construction. \cite{huang2019towards} develop a KG centered on healthy eating and TCM concepts. Data from three major nutrition-related websites are processed using NLP and machine learning techniques to extract entities such as foods, symptoms, populations, and nutrients. Relationships including good-for, bad-for, part-of, contain, and related are identified and encoded. The method uses CRF for entity recognition, SVM for relation classification, and TF-IDF combined with decision trees for entity similarity measurement, ensuring that both entities and relations reflect scientifically grounded nutritional and TCM knowledge. The resulting KG can support health-aware recommendation, semantic search, and intelligent QA, and serves as an infrastructural component for smart dietary management systems.

Overall, the reviewed work on KG generation highlights several common design themes. High-quality medical KGs typically emerge from pipelines that: (i) integrate heterogeneous textual and structured sources (EHRs, guidelines, literature, web content); (ii) combine statistical and deep learning methods for entity and relation extraction with explicit ontologies and expert validation; (iii) incorporate mechanisms for normalization, synonym handling, and ontology enrichment to maintain semantic coherence; and (iv) use pruning or clustering strategies to manage noise and focus on clinically meaningful subgraphs. Despite substantial progress, open challenges remain in scaling these frameworks to continuously updated data streams, ensuring robustness under distribution shifts, encoding causal and temporal structure in a principled way, and aligning KG construction choices with the requirements of downstream applications such as decision support, prediction, and QA.

\begin{figure*}[t]
\centering
\includegraphics[width=.9\textwidth]{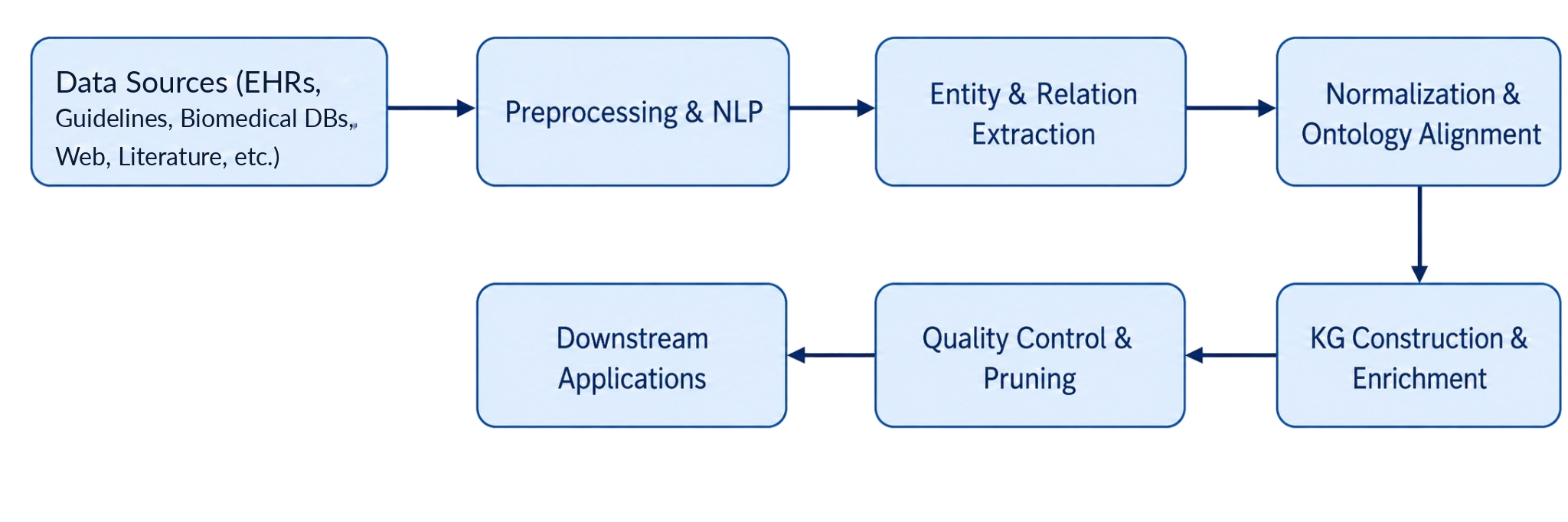}
\caption{Generic pipeline for constructing medical knowledge graphs from heterogeneous data sources. Raw data from EHRs, clinical guidelines, biomedical databases, and web or literature resources are first preprocessed and processed with NLP techniques for entity and relation extraction. Extracted concepts are then normalized and aligned with clinical ontologies, before being assembled into a knowledge graph, refined through quality control and pruning, and finally used in downstream applications such as decision support, prediction, recommendation, and QA.}\label{fig:fig6}
\end{figure*}

\section{Discussion}
\label{sec:sec4}

With the increasing complexity and heterogeneity of biomedical data, the growing emphasis on precision medicine, and the rising demand for AI-driven healthcare systems, there is a pressing need for frameworks that can represent, integrate, and reason over multimodal and multidimensional knowledge. Knowledge graphs (KGs) offer a compelling paradigm for this purpose: they encode medical entities and their relationships in a structured, semantically rich form, providing a common backbone on which analytical, predictive, and decision-support models can be built.

This survey has provided a structured overview of how KGs are being used across five major areas of healthcare AI—clinical decision support, prediction, recommender systems, precision medicine, and medical question answering—and has also examined methodological work on medical KG generation, from EHR-derived graphs and semantic web–based frameworks to deep learning–driven extraction pipelines. Across these application domains, several cross-cutting themes emerge.
First, KGs play a central role in explainability and transparency. In clinical decision support and prediction tasks, graph-based representations enable models to expose risk factors, causal chains, or meta-paths that connect patient attributes, biomarkers, treatments, and outcomes. Methods that integrate GNNs with medical KGs can produce patient-level embeddings while still allowing practitioners to trace which nodes, edges, or subgraphs contributed most to a prediction, thereby supporting human oversight rather than replacing it. In lifestyle and dietary recommendation, KGs connect user preferences and constraints (e.g., allergies, chronic conditions) to nutritional knowledge and disease guidelines, making it possible to justify recommendations in terms of both taste similarity and health considerations.
Second, KGs facilitate personalization by linking population-level knowledge to individual-level data. Personalized or patient-specific KGs (e.g., PHKGs, iKGs, PCKGs) combine EHRs, genomic or molecular data, lifestyle factors, and guideline-derived knowledge into unified, individual graphs. These graphs can then support tailored risk prediction, medication selection, and treatment planning, as well as digital twin simulations that anticipate disease trajectories or treatment responses. In QA systems, KGs enable case-based or patient-context–aware answering that goes beyond generic medical advice.

Third, KGs provide a natural substrate for hybrid and neuro-symbolic AI, where symbolic constraints, ontologies, and causal structure are combined with statistical learning and, more recently, LLMs. Hybrid frameworks can use KGs to constrain or ground generative models, to encode SHACL constraints and Horn rules for valid link prediction, or to support counterfactual reasoning over treatment options. In QA and dialogue settings, KG-augmented RAG architectures help control hallucinations and keep generated answers anchored in trusted biomedical knowledge.
At the same time, the survey of KG construction frameworks underscores that downstream performance and interpretability depend heavily on how graphs are built. Choices about data sources, ontologies, entity and relation extraction methods, normalization, and pruning directly impact graph coverage, noise levels, and suitability for specific clinical tasks. Designing reusable, robust KG generation pipelines is therefore as important as designing downstream models.

Despite these advances, medical KGs still face substantial limitations that prevent their full potential from being realized in routine practice. The following sections summarize key shortcomings and outline open research directions.

\subsection{Limitations and Shortcomings of Medical Knowledge Graphs}
\label{subsec1}

Despite remarkable progress, current medical KGs and KG-based systems exhibit several important limitations.
A first overarching challenge is limited and fragmented coverage. Many existing KGs focus on narrow subdomains—such as a specific disease area, data modality, or type of interaction—or are built from a small number of curated sources. This domain specificity can be advantageous for focused tasks but restricts generalization to broader clinical contexts. When KGs built for one condition, population, or institution are applied elsewhere, missing entities, relations, or context can lead to degraded performance and brittle behavior.
Closely related is the problem of heterogeneous data integration and interoperability. Biomedical information is spread across EHR systems, registries, clinical guidelines, ontologies, omics databases, and literature, each with its own standards, formats, and terminologies. Aligning these sources—resolving different coding schemes, handling synonyms and ambiguities, and managing conflicting or outdated information—remains a major obstacle \cite{wu2023medical}. While mapping to common vocabularies and ontologies can mitigate some issues, the resulting KGs often inherit inconsistencies, gaps, and biases from their source data.
A third limitation concerns reasoning algorithms and representation learning. Many works rely on embedding-based models for link prediction or node classification, or on relatively simple rule-based reasoning. However, these methods can be fragile when applied to dense, multi-relational biomedical graphs. Embedding models may struggle to represent complex, higher-order dependencies, and their predictions can be difficult to interpret without additional tooling. Symbolic or rule-based reasoners, on the other hand, may lack robustness and scalability when confronted with noisy, incomplete, or evolving graphs. Hybrid approaches can help, but they introduce new design choices about how to combine symbolic and numerical components in a stable and trustworthy way.

Text-centric KG construction adds further difficulties. In specialized domains such as oncology, entity and relation extraction is complicated by nested entities, long and compound expressions, and domain-specific jargon. Pipeline-based architectures, where entity recognition and relation extraction are performed in sequence, are especially vulnerable to cascading errors: mistakes in early stages propagate and amplify in downstream graph construction and reasoning \cite{yang2023comprehensive}. Deep learning–based extraction frameworks (e.g., involving BioBERT or other domain-specific models) can improve recall and precision, but they still require extensive annotation, careful evaluation, and mechanisms to handle uncertainty and conflicting evidence.

Beyond technical aspects, ethical and privacy considerations pose persistent challenges. KGs that directly encode patient-level data or are exposed to end users through QA systems and recommender interfaces raise concerns about confidentiality, re-identification risks, and unintended inference of sensitive attributes. When KGs are used to guide clinical decisions, recommendation or triage, issues of bias, fairness, and accountability also arise: biased training data or incomplete graphs can systematically disadvantage certain groups, and opaque reasoning pipelines make it difficult to assign responsibility for errors. These concerns are exacerbated in interactive applications where users may over-trust AI outputs simply because they appear “knowledge-based.”
Taken together, these limitations highlight the need for medical KG frameworks that are more comprehensive, interoperable, robust, and ethically grounded—from the way graphs are constructed and updated to the algorithms that operate on them and the interfaces through which their outputs are delivered.

\subsection{Open Problems and Future Directions}
\label{subsec1}
Against this backdrop, several research avenues remain open and appear particularly promising for advancing KG-based healthcare systems.
First, there is a clear need for dynamic, patient-centric knowledge graphs that evolve with an individual’s medical history, demographics, comorbidities, and treatment context. Rather than static, institution-level graphs, future systems should support real-time or near–real-time updates from EHRs, wearable devices, and other data streams, while preserving temporal and causal structure. Such personalized KGs could underpin digital twins and individualized decision support, but they raise open questions about efficient incremental updates, temporal reasoning, and safe deployment in clinical workflows.

Second, the integration of large language models (LLMs) with biomedical KGs remains an active research frontier. While current KG–LLM hybrids demonstrate promising results in QA, dialogue, and KG construction, there is still limited understanding of how best to orchestrate the interaction between free-text reasoning and structured knowledge. Future work may explore architectures in which LLMs help extract, normalize, and update KG content, while KGs constrain and ground LLM outputs; investigate methods for learning to retrieve, rank, and verbalize relevant subgraphs in a task-aware manner; and develop protocols for auditing how LLMs use KG context to ensure that generated answers remain faithful and non-hallucinated.
Third, advancing trustworthy and interpretable inference algorithms remains crucial. Beyond improving accuracy, models operating on medical KGs should provide transparent reasoning paths, confidence estimates, and human-understandable explanations. This includes work on explainable graph neural networks, causal discovery and counterfactual reasoning over KGs, and hybrid neuro-symbolic systems that can expose the rules or constraints driving their predictions. Ensuring robustness under distribution shifts, noisy or incomplete graphs, and adversarial perturbations is equally important for real-world adoption.

Fourth, future KGs are likely to be increasingly multimodal. Integrating clinical text, structured EHRs, medical imaging, physiological time series, genomic and proteomic profiles, and sensor data into unified knowledge representations could unlock richer analyses and more holistic models of health and disease. This raises both representational challenges—how to encode images or signals within graph structures—and algorithmic ones, including multimodal fusion, cross-modal reasoning, and scalable training on large, heterogeneous datasets.
Fifth, there is a need for more systematic work on KG construction pipelines and governance in medicine. This includes methods for continuous graph maintenance, provenance tracking, handling conflicting evidence, and encoding versioned or context-dependent knowledge (e.g., changing guidelines, evolving disease definitions). Closer coupling between KG construction frameworks and downstream tasks—such as tailoring extraction and normalization strategies to the needs of CDSS, prediction, or QA systems—could help reduce mismatch between graph design and application requirements.

Finally, the field would benefit from specialized evaluation metrics and benchmarks for KG-based healthcare models. Existing metrics often focus narrowly on link prediction accuracy or downstream classification performance. Future benchmarks should account for interpretability (e.g., quality and usefulness of explanations), reasoning consistency, knowledge coverage and completeness, fairness across patient subgroups, and safety properties relevant to clinical contexts. Shared datasets and challenge tasks that combine KG construction, reasoning, and application-level evaluation would foster more rigorous comparisons and accelerate progress.
Overall, addressing these open problems will require close collaboration between AI researchers, clinicians, domain experts, and ethicists, as well as careful consideration of regulatory and organizational constraints in healthcare systems.

\section{Conclusion}
\label{sec:sec5}
In this paper, we conducted a structured review of how knowledge graphs are used in the medical domain, with a dual focus on application areas and KG generation methodologies. On the application side, we organized existing work into five major categories—clinical decision support, prediction, health recommendation, precision medicine, and medical question answering—and examined how clinical and biomedical data are encoded as graphs, which algorithms are employed (e.g., KG embeddings, GNNs, hybrid neuro-symbolic models, KG–LLM architectures), and what specific clinical or operational goals these systems pursue. On the construction side, we surveyed representative frameworks for building medical KGs from EHRs, clinical narratives, biomedical literature, and web resources, highlighting common design patterns for entity and relation extraction, ontology alignment, and graph enrichment.

Across these strands of work, our analysis indicates that KGs can substantially enhance interpretability, semantic coherence, and trustworthiness in medical AI systems by providing explicit structure and a shared vocabulary over which reasoning and learning can operate. At the same time, we identified key limitations, including restricted and fragmented coverage of existing graphs, challenges in aligning heterogeneous data sources, the fragility of current reasoning and representation-learning methods on dense multi-relational graphs, and unresolved issues around privacy, bias, and accountability.

By synthesizing both application-driven and methodology-oriented research, we hope this survey can serve as a useful reference for researchers and practitioners designing KG-based intelligent systems in healthcare. In particular, the open problems and future directions outlined above—ranging from dynamic, patient-centric KGs and KG–LLM integration to multimodal graph representations and domain-specific evaluation frameworks—offer concrete avenues for advancing the role of knowledge graphs in safe, effective, and equitable AI-driven healthcare.





\nocite{*}
\begin{thebibliography}{86}
	\expandafter\ifx\csname natexlab\endcsname\relax\def\natexlab#1{#1}\fi
	\providecommand{\url}[1]{\texttt{#1}}
	\providecommand{\href}[2]{#2}
	\providecommand{\path}[1]{#1}
	\providecommand{\DOIprefix}{doi:}
	\providecommand{\ArXivprefix}{arXiv:}
	\providecommand{\URLprefix}{URL: }
	\providecommand{\Pubmedprefix}{pmid:}
	\providecommand{\doi}[1]{\href{http://dx.doi.org/#1}{\path{#1}}}
	\providecommand{\Pubmed}[1]{\href{pmid:#1}{\path{#1}}}
	\providecommand{\bibinfo}[2]{#2}
	\ifx\xfnm\relax \def\xfnm[#1]{\unskip,\space#1}\fi
	\bibitem[{Alam et~al.(2023)Alam, Giglou and Malik}]{alam2023automated}
	\bibinfo{author}{Alam, F.}, \bibinfo{author}{Giglou, H.B.},
	\bibinfo{author}{Malik, K.M.}, \bibinfo{year}{2023}.
	\newblock \bibinfo{title}{Automated clinical knowledge graph generation
		framework for evidence based medicine}.
	\newblock \bibinfo{journal}{Expert systems with applications}
	\bibinfo{volume}{233}, \bibinfo{pages}{120964}.
	\bibitem[{Bhoi et~al.(2021)Bhoi, Lee, Hsu, Fang and
		Tan}]{bhoi2021personalizing}
	\bibinfo{author}{Bhoi, S.}, \bibinfo{author}{Lee, M.L.}, \bibinfo{author}{Hsu,
		W.}, \bibinfo{author}{Fang, H.S.A.}, \bibinfo{author}{Tan, N.C.},
	\bibinfo{year}{2021}.
	\newblock \bibinfo{title}{Personalizing medication recommendation with a
		graph-based approach}.
	\newblock \bibinfo{journal}{ACM Transactions on Information Systems (TOIS)}
	\bibinfo{volume}{40}, \bibinfo{pages}{1--23}.
	\bibitem[{Chandak et~al.(2023)Chandak, Huang and Zitnik}]{chandak2023building}
	\bibinfo{author}{Chandak, P.}, \bibinfo{author}{Huang, K.},
	\bibinfo{author}{Zitnik, M.}, \bibinfo{year}{2023}.
	\newblock \bibinfo{title}{Building a knowledge graph to enable precision
		medicine}.
	\newblock \bibinfo{journal}{Scientific data} \bibinfo{volume}{10},
	\bibinfo{pages}{67}.
	\bibitem[{Chaoyu and Lei(2020)}]{chaoyu2020review}
	\bibinfo{author}{Chaoyu, Z.}, \bibinfo{author}{Lei, L.}, \bibinfo{year}{2020}.
	\newblock \bibinfo{title}{A review of medical decision supports based on
		knowledge graph}.
	\newblock \bibinfo{journal}{Data Anal Knowl Discovery} \bibinfo{volume}{4},
	\bibinfo{pages}{26--32}.
	\bibitem[{Che et~al.(2021)Che, Yao, Che, Cao and Kong}]{che2021knowledge}
	\bibinfo{author}{Che, M.}, \bibinfo{author}{Yao, K.}, \bibinfo{author}{Che,
		C.}, \bibinfo{author}{Cao, Z.}, \bibinfo{author}{Kong, F.},
	\bibinfo{year}{2021}.
	\newblock \bibinfo{title}{Knowledge-graph-based drug repositioning against
		covid-19 by graph convolutional network with attention mechanism}.
	\newblock \bibinfo{journal}{Future Internet} \bibinfo{volume}{13},
	\bibinfo{pages}{13}.
	\bibitem[{Chen et~al.(2019)Chen, Agrawal, Horng and Sontag}]{chen2019robustly}
	\bibinfo{author}{Chen, I.Y.}, \bibinfo{author}{Agrawal, M.},
	\bibinfo{author}{Horng, S.}, \bibinfo{author}{Sontag, D.},
	\bibinfo{year}{2019}.
	\newblock \bibinfo{title}{Robustly extracting medical knowledge from ehrs: a
		case study of learning a health knowledge graph}, in:
	\bibinfo{booktitle}{Pacific Symposium on Biocomputing 2020},
	\bibinfo{organization}{World Scientific}. pp. \bibinfo{pages}{19--30}.
	\bibitem[{Chen et~al.(2021)Chen, Subburathinam, Chen and
		Zaki}]{chen2021personalized}
	\bibinfo{author}{Chen, Y.}, \bibinfo{author}{Subburathinam, A.},
	\bibinfo{author}{Chen, C.H.}, \bibinfo{author}{Zaki, M.J.},
	\bibinfo{year}{2021}.
	\newblock \bibinfo{title}{Personalized food recommendation as constrained
		question answering over a large-scale food knowledge graph}, in:
	\bibinfo{booktitle}{Proceedings of the 14th ACM international conference on
		web search and data mining}, pp. \bibinfo{pages}{544--552}.
	\bibitem[{Chi et~al.(2018)Chi, Yu, Qi and Xu}]{chi2018knowledge}
	\bibinfo{author}{Chi, Y.}, \bibinfo{author}{Yu, C.}, \bibinfo{author}{Qi, X.},
	\bibinfo{author}{Xu, H.}, \bibinfo{year}{2018}.
	\newblock \bibinfo{title}{Knowledge management in healthcare sustainability: a
		smart healthy diet assistant in traditional chinese medicine culture}.
	\newblock \bibinfo{journal}{Sustainability} \bibinfo{volume}{10},
	\bibinfo{pages}{4197}.
	\bibitem[{Choi and Lee(2021)}]{choi2021identifying}
	\bibinfo{author}{Choi, W.}, \bibinfo{author}{Lee, H.}, \bibinfo{year}{2021}.
	\newblock \bibinfo{title}{Identifying disease-gene associations using a
		convolutional neural network-based model by embedding a biological knowledge
		graph with entity descriptions}.
	\newblock \bibinfo{journal}{Plos one} \bibinfo{volume}{16},
	\bibinfo{pages}{e0258626}.
	\bibitem[{Duan et~al.(2025)Duan, Zhou, Li, Qin, Wang, Kan and
		Hu}]{duan2025research}
	\bibinfo{author}{Duan, Y.}, \bibinfo{author}{Zhou, Q.}, \bibinfo{author}{Li,
		Y.}, \bibinfo{author}{Qin, C.}, \bibinfo{author}{Wang, Z.},
	\bibinfo{author}{Kan, H.}, \bibinfo{author}{Hu, J.}, \bibinfo{year}{2025}.
	\newblock \bibinfo{title}{Research on a traditional chinese medicine case-based
		question-answering system integrating large language models and knowledge
		graphs}.
	\newblock \bibinfo{journal}{Frontiers in Medicine} \bibinfo{volume}{11},
	\bibinfo{pages}{1512329}.
	\bibitem[{Elahi et~al.(2024)Elahi, Anwar, Shah, Halim, Ullah, Rida and
		Waqas}]{elahi2024knowledge}
	\bibinfo{author}{Elahi, E.}, \bibinfo{author}{Anwar, S.},
	\bibinfo{author}{Shah, B.}, \bibinfo{author}{Halim, Z.},
	\bibinfo{author}{Ullah, A.}, \bibinfo{author}{Rida, I.},
	\bibinfo{author}{Waqas, M.}, \bibinfo{year}{2024}.
	\newblock \bibinfo{title}{Knowledge graph enhanced contextualized
		attention-based network for responsible user-specific recommendation}.
	\newblock \bibinfo{journal}{ACM Transactions on Intelligent Systems and
		Technology} \bibinfo{volume}{15}, \bibinfo{pages}{1--24}.
	\bibitem[{Freidel and Schwarz(2025)}]{freidel2025knowledge}
	\bibinfo{author}{Freidel, S.}, \bibinfo{author}{Schwarz, E.},
	\bibinfo{year}{2025}.
	\newblock \bibinfo{title}{Knowledge graphs in psychiatric research: Potential
		applications and future perspectives}.
	\newblock \bibinfo{journal}{Acta Psychiatrica Scandinavica}
	\bibinfo{volume}{151}, \bibinfo{pages}{180--191}.
	\bibitem[{Gao et~al.(2023)Gao, Yin, Wang, Wang, Du and Li}]{gao2023medical}
	\bibinfo{author}{Gao, C.}, \bibinfo{author}{Yin, S.}, \bibinfo{author}{Wang,
		H.}, \bibinfo{author}{Wang, Z.}, \bibinfo{author}{Du, Z.},
	\bibinfo{author}{Li, X.}, \bibinfo{year}{2023}.
	\newblock \bibinfo{title}{Medical-knowledge-based graph neural network for
		medication combination prediction}.
	\newblock \bibinfo{journal}{IEEE Transactions on Neural Networks and Learning
		Systems} \bibinfo{volume}{35}, \bibinfo{pages}{13246--13257}.
	\bibitem[{Gao et~al.(2025)Gao, Li, Croxford, Caskey, Patterson, Churpek,
		Miller, Dligach and Afshar}]{gao2025leveraging}
	\bibinfo{author}{Gao, Y.}, \bibinfo{author}{Li, R.}, \bibinfo{author}{Croxford,
		E.}, \bibinfo{author}{Caskey, J.}, \bibinfo{author}{Patterson, B.W.},
	\bibinfo{author}{Churpek, M.}, \bibinfo{author}{Miller, T.},
	\bibinfo{author}{Dligach, D.}, \bibinfo{author}{Afshar, M.},
	\bibinfo{year}{2025}.
	\newblock \bibinfo{title}{Leveraging medical knowledge graphs into large
		language models for diagnosis prediction: Design and application study}.
	\newblock \bibinfo{journal}{Jmir Ai} \bibinfo{volume}{4},
	\bibinfo{pages}{e58670}.
	\bibitem[{Gunjan et~al.(2022)Gunjan, Vijayalata, Valli, Kumar, Mohamed and
		Saravanan}]{gunjan2022retracted}
	\bibinfo{author}{Gunjan, V.K.}, \bibinfo{author}{Vijayalata, Y.},
	\bibinfo{author}{Valli, S.}, \bibinfo{author}{Kumar, S.},
	\bibinfo{author}{Mohamed, M.}, \bibinfo{author}{Saravanan, V.},
	\bibinfo{year}{2022}.
	\newblock \bibinfo{title}{[retracted] machine learning and cloud-based
		knowledge graphs to recognize suicidal mental tendencies}.
	\newblock \bibinfo{journal}{Computational Intelligence and Neuroscience}
	\bibinfo{volume}{2022}, \bibinfo{pages}{3604113}.
	\bibitem[{Gyrard et~al.(2022)Gyrard, Jaimini, Gaur, Shekharpour, Thirunarayan
		and Sheth}]{gyrard2022reasoning}
	\bibinfo{author}{Gyrard, A.}, \bibinfo{author}{Jaimini, U.},
	\bibinfo{author}{Gaur, M.}, \bibinfo{author}{Shekharpour, S.},
	\bibinfo{author}{Thirunarayan, K.}, \bibinfo{author}{Sheth, A.},
	\bibinfo{year}{2022}.
	\newblock \bibinfo{title}{Reasoning over personalized healthcare knowledge
		graph: a case study of patients with allergies and symptoms}, in:
	\bibinfo{booktitle}{Semantic Models in IoT and Ehealth Applications}.
	\bibinfo{publisher}{Elsevier}, pp. \bibinfo{pages}{199--225}.
	\bibitem[{Hamane et~al.(2023)Hamane, Samih and
		Fennan}]{hamane2023healthpathfinder}
	\bibinfo{author}{Hamane, Z.}, \bibinfo{author}{Samih, A.},
	\bibinfo{author}{Fennan, A.}, \bibinfo{year}{2023}.
	\newblock \bibinfo{title}{Healthpathfinder: Navigating the healthcare knowledge
		graph with neural attention for personalized health recommendations}, in:
	\bibinfo{booktitle}{The Proceedings of the International Conference on Smart
		City Applications}, \bibinfo{organization}{Springer}. pp.
	\bibinfo{pages}{429--446}.
	\bibitem[{Hao et~al.(2023)Hao, Chen, Pan, Qiu, Zhang, Yu, Han and
		Du}]{hao2023enhancing}
	\bibinfo{author}{Hao, X.}, \bibinfo{author}{Chen, Q.}, \bibinfo{author}{Pan,
		H.}, \bibinfo{author}{Qiu, J.}, \bibinfo{author}{Zhang, Y.},
	\bibinfo{author}{Yu, Q.}, \bibinfo{author}{Han, Z.}, \bibinfo{author}{Du,
		X.}, \bibinfo{year}{2023}.
	\newblock \bibinfo{title}{Enhancing drug--drug interaction prediction by
		three-way decision and knowledge graph embedding}.
	\newblock \bibinfo{journal}{Granular Computing} \bibinfo{volume}{8},
	\bibinfo{pages}{67--76}.
	\bibitem[{Huang et~al.(2022)Huang, Shi, Wang, Wang and Han}]{huang2022novel}
	\bibinfo{author}{Huang, B.}, \bibinfo{author}{Shi, X.}, \bibinfo{author}{Wang,
		R.}, \bibinfo{author}{Wang, C.}, \bibinfo{author}{Han, Y.},
	\bibinfo{year}{2022}.
	\newblock \bibinfo{title}{A novel recipes recommendation system based on
		knowledge-graph}, in: \bibinfo{booktitle}{2022 7th International Conference
		on Intelligent Computing and Signal Processing (ICSP)},
	\bibinfo{organization}{IEEE}. pp. \bibinfo{pages}{1408--1412}.
	\bibitem[{Huang et~al.(2019)Huang, Yu, Chi, Qi and Xu}]{huang2019towards}
	\bibinfo{author}{Huang, L.}, \bibinfo{author}{Yu, C.}, \bibinfo{author}{Chi,
		Y.}, \bibinfo{author}{Qi, X.}, \bibinfo{author}{Xu, H.},
	\bibinfo{year}{2019}.
	\newblock \bibinfo{title}{Towards smart healthcare management based on
		knowledge graph technology}, in: \bibinfo{booktitle}{Proceedings of the 2019
		8th International Conference on Software and Computer Applications}, pp.
	\bibinfo{pages}{330--337}.
	\bibitem[{Huang et~al.(2024)Huang, Zhang, Bhatti, Wu, Zhang and
		Ghadi}]{huang2024interpretable}
	\bibinfo{author}{Huang, M.}, \bibinfo{author}{Zhang, X.S.},
	\bibinfo{author}{Bhatti, U.A.}, \bibinfo{author}{Wu, Y.},
	\bibinfo{author}{Zhang, Y.}, \bibinfo{author}{Ghadi, Y.Y.},
	\bibinfo{year}{2024}.
	\newblock \bibinfo{title}{An interpretable approach using hybrid graph networks
		and explainable ai for intelligent diagnosis recommendations in chronic
		disease care}.
	\newblock \bibinfo{journal}{Biomedical Signal Processing and Control}
	\bibinfo{volume}{91}, \bibinfo{pages}{105913}.
	\bibitem[{Jamrat et~al.(2023)Jamrat, Sukasem, Sratthaphut, Hongkaew and
		Samanchuen}]{jamrat2023precision}
	\bibinfo{author}{Jamrat, S.}, \bibinfo{author}{Sukasem, C.},
	\bibinfo{author}{Sratthaphut, L.}, \bibinfo{author}{Hongkaew, Y.},
	\bibinfo{author}{Samanchuen, T.}, \bibinfo{year}{2023}.
	\newblock \bibinfo{title}{A precision medicine approach to personalized
		prescribing using genetic and nongenetic factors for clinical
		decision-making}.
	\newblock \bibinfo{journal}{Computers in Biology and Medicine}
	\bibinfo{volume}{165}, \bibinfo{pages}{107329}.
	\bibitem[{Kr{\"a}mer et~al.(2021)Kr{\"a}mer, Billaud, Tugendreich, Shiffman,
		Jones and Green}]{kramer2021coronavirus}
	\bibinfo{author}{Kr{\"a}mer, A.}, \bibinfo{author}{Billaud, J.N.},
	\bibinfo{author}{Tugendreich, S.}, \bibinfo{author}{Shiffman, D.},
	\bibinfo{author}{Jones, M.}, \bibinfo{author}{Green, J.},
	\bibinfo{year}{2021}.
	\newblock \bibinfo{title}{The coronavirus network explorer: Mining a
		large-scale knowledge graph for effects of sars-cov-2 on host cell function}.
	\newblock \bibinfo{journal}{BMC bioinformatics} \bibinfo{volume}{22},
	\bibinfo{pages}{229}.
	\bibitem[{Lan et~al.(2022)Lan, Dong, Chen, Zheng, Liu, Pan and
		Chen}]{lan2022kgancda}
	\bibinfo{author}{Lan, W.}, \bibinfo{author}{Dong, Y.}, \bibinfo{author}{Chen,
		Q.}, \bibinfo{author}{Zheng, R.}, \bibinfo{author}{Liu, J.},
	\bibinfo{author}{Pan, Y.}, \bibinfo{author}{Chen, Y.P.P.},
	\bibinfo{year}{2022}.
	\newblock \bibinfo{title}{Kgancda: predicting circrna-disease associations
		based on knowledge graph attention network}.
	\newblock \bibinfo{journal}{Briefings in Bioinformatics} \bibinfo{volume}{23},
	\bibinfo{pages}{bbab494}.
	\bibitem[{Li et~al.(2023a)Li, Zaki and Chen}]{li2023health}
	\bibinfo{author}{Li, D.}, \bibinfo{author}{Zaki, M.J.}, \bibinfo{author}{Chen,
		C.h.}, \bibinfo{year}{2023}a.
	\newblock \bibinfo{title}{Health-guided recipe recommendation over knowledge
		graphs}.
	\newblock \bibinfo{journal}{Journal of Web Semantics} \bibinfo{volume}{75},
	\bibinfo{pages}{100743}.
	\bibitem[{Li et~al.(2022)Li, Zhong, Yang, Duan, Wang, Wu and He}]{li2022deepkg}
	\bibinfo{author}{Li, Z.}, \bibinfo{author}{Zhong, Q.}, \bibinfo{author}{Yang,
		J.}, \bibinfo{author}{Duan, Y.}, \bibinfo{author}{Wang, W.},
	\bibinfo{author}{Wu, C.}, \bibinfo{author}{He, K.}, \bibinfo{year}{2022}.
	\newblock \bibinfo{title}{Deepkg: an end-to-end deep learning-based workflow
		for biomedical knowledge graph extraction, optimization and applications}.
	\newblock \bibinfo{journal}{Bioinformatics} \bibinfo{volume}{38},
	\bibinfo{pages}{1477--1479}.
	\bibitem[{Li et~al.(2023b)Li, Fu, Li, Fan, Li, Wang and
		Zhou}]{li2023prediction}
	\bibinfo{author}{Li, Z.Q.}, \bibinfo{author}{Fu, Z.X.}, \bibinfo{author}{Li,
		W.J.}, \bibinfo{author}{Fan, H.}, \bibinfo{author}{Li, S.N.},
	\bibinfo{author}{Wang, X.M.}, \bibinfo{author}{Zhou, P.},
	\bibinfo{year}{2023}b.
	\newblock \bibinfo{title}{Prediction of diabetic macular edema using knowledge
		graph}.
	\newblock \bibinfo{journal}{Diagnostics} \bibinfo{volume}{13},
	\bibinfo{pages}{1858}.
	\bibitem[{Lin et~al.(2020)Lin, Yang and Yin}]{lin2020patient}
	\bibinfo{author}{Lin, Z.}, \bibinfo{author}{Yang, D.}, \bibinfo{author}{Yin,
		X.}, \bibinfo{year}{2020}.
	\newblock \bibinfo{title}{Patient similarity via joint embeddings of medical
		knowledge graph and medical entity descriptions}.
	\newblock \bibinfo{journal}{IEEE Access} \bibinfo{volume}{8},
	\bibinfo{pages}{156663--156676}.
	\bibitem[{Liu et~al.(2021)Liu, Pan, Wang, Feenstra, Heringa and
		Huang}]{liu2021exploring}
	\bibinfo{author}{Liu, T.}, \bibinfo{author}{Pan, X.}, \bibinfo{author}{Wang,
		X.}, \bibinfo{author}{Feenstra, K.A.}, \bibinfo{author}{Heringa, J.},
	\bibinfo{author}{Huang, Z.}, \bibinfo{year}{2021}.
	\newblock \bibinfo{title}{Exploring the microbiota-gut-brain axis for mental
		disorders with knowledge graphs}.
	\newblock \bibinfo{journal}{Journal of Artificial Intelligence for Medical
		Sciences} \bibinfo{volume}{1}, \bibinfo{pages}{30--42}.
	\bibitem[{Lu and Uddin(2023)}]{lu2023disease}
	\bibinfo{author}{Lu, H.}, \bibinfo{author}{Uddin, S.}, \bibinfo{year}{2023}.
	\newblock \bibinfo{title}{Disease prediction using graph machine learning based
		on electronic health data: a review of approaches and trends}, in:
	\bibinfo{booktitle}{Healthcare}, \bibinfo{organization}{MDPI}. p.
	\bibinfo{pages}{1031}.
	\bibitem[{Lyu et~al.(2023)Lyu, Tian, Shang, Zhou, Yang, Liu, Yao, Zhang, Chen
		and Li}]{lyu2023causal}
	\bibinfo{author}{Lyu, K.}, \bibinfo{author}{Tian, Y.}, \bibinfo{author}{Shang,
		Y.}, \bibinfo{author}{Zhou, T.}, \bibinfo{author}{Yang, Z.},
	\bibinfo{author}{Liu, Q.}, \bibinfo{author}{Yao, X.}, \bibinfo{author}{Zhang,
		P.}, \bibinfo{author}{Chen, J.}, \bibinfo{author}{Li, J.},
	\bibinfo{year}{2023}.
	\newblock \bibinfo{title}{Causal knowledge graph construction and evaluation
		for clinical decision support of diabetic nephropathy}.
	\newblock \bibinfo{journal}{Journal of Biomedical Informatics}
	\bibinfo{volume}{139}, \bibinfo{pages}{104298}.
	\bibitem[{Ma et~al.(2023)Ma, Zhou, Liu and Koslicki}]{ma2023kgml}
	\bibinfo{author}{Ma, C.}, \bibinfo{author}{Zhou, Z.}, \bibinfo{author}{Liu,
		H.}, \bibinfo{author}{Koslicki, D.}, \bibinfo{year}{2023}.
	\newblock \bibinfo{title}{Kgml-xdtd: a knowledge graph--based machine learning
		framework for drug treatment prediction and mechanism description}.
	\newblock \bibinfo{journal}{GigaScience} \bibinfo{volume}{12},
	\bibinfo{pages}{giad057}.
	\bibitem[{Ma et~al.(2018)Ma, You, Xiao, Chitta, Zhou and Gao}]{ma2018kame}
	\bibinfo{author}{Ma, F.}, \bibinfo{author}{You, Q.}, \bibinfo{author}{Xiao,
		H.}, \bibinfo{author}{Chitta, R.}, \bibinfo{author}{Zhou, J.},
	\bibinfo{author}{Gao, J.}, \bibinfo{year}{2018}.
	\newblock \bibinfo{title}{Kame: Knowledge-based attention model for diagnosis
		prediction in healthcare}, in: \bibinfo{booktitle}{Proceedings of the 27th
		ACM international conference on information and knowledge management}, pp.
	\bibinfo{pages}{743--752}.
	\bibitem[{Malik et~al.(2020)Malik, Krishnamurthy, Alobaidi, Hussain, Alam and
		Malik}]{malik2020automated}
	\bibinfo{author}{Malik, K.M.}, \bibinfo{author}{Krishnamurthy, M.},
	\bibinfo{author}{Alobaidi, M.}, \bibinfo{author}{Hussain, M.},
	\bibinfo{author}{Alam, F.}, \bibinfo{author}{Malik, G.},
	\bibinfo{year}{2020}.
	\newblock \bibinfo{title}{Automated domain-specific healthcare knowledge graph
		curation framework: Subarachnoid hemorrhage as phenotype}.
	\newblock \bibinfo{journal}{Expert Systems with Applications}
	\bibinfo{volume}{145}, \bibinfo{pages}{113120}.
	\bibitem[{Mohamed et~al.(2020)Mohamed, Nov{\'a}{\v{c}}ek and
		Nounu}]{mohamed2020discovering}
	\bibinfo{author}{Mohamed, S.K.}, \bibinfo{author}{Nov{\'a}{\v{c}}ek, V.},
	\bibinfo{author}{Nounu, A.}, \bibinfo{year}{2020}.
	\newblock \bibinfo{title}{Discovering protein drug targets using knowledge
		graph embeddings}.
	\newblock \bibinfo{journal}{Bioinformatics} \bibinfo{volume}{36},
	\bibinfo{pages}{603--610}.
	\bibitem[{Mohammadhassanzadeh et~al.(2018)Mohammadhassanzadeh, Abidi,
		Van~Woensel and Abidi}]{mohammadhassanzadeh2018investigating}
	\bibinfo{author}{Mohammadhassanzadeh, H.}, \bibinfo{author}{Abidi, S.R.},
	\bibinfo{author}{Van~Woensel, W.}, \bibinfo{author}{Abidi, S.S.R.},
	\bibinfo{year}{2018}.
	\newblock \bibinfo{title}{Investigating plausible reasoning over knowledge
		graphs for semantics-based health data analytics}, in:
	\bibinfo{booktitle}{2018 IEEE 27th International Conference on Enabling
		Technologies: Infrastructure for Collaborative Enterprises (WETICE)},
	\bibinfo{organization}{IEEE}. pp. \bibinfo{pages}{148--153}.
	\bibitem[{Morris et~al.(2023)Morris, Soman, Akbas, Zhou, Smith, Meng, Huang,
		Cerono, Schenk, Rizk-Jackson et~al.}]{morris2023scalable}
	\bibinfo{author}{Morris, J.H.}, \bibinfo{author}{Soman, K.},
	\bibinfo{author}{Akbas, R.E.}, \bibinfo{author}{Zhou, X.},
	\bibinfo{author}{Smith, B.}, \bibinfo{author}{Meng, E.C.},
	\bibinfo{author}{Huang, C.C.}, \bibinfo{author}{Cerono, G.},
	\bibinfo{author}{Schenk, G.}, \bibinfo{author}{Rizk-Jackson, A.}, et~al.,
	\bibinfo{year}{2023}.
	\newblock \bibinfo{title}{The scalable precision medicine open knowledge engine
		(spoke): a massive knowledge graph of biomedical information}.
	\newblock \bibinfo{journal}{Bioinformatics} \bibinfo{volume}{39},
	\bibinfo{pages}{btad080}.
	\bibitem[{Mu{\~n}oz et~al.(2019)Mu{\~n}oz, Nov{\'a}{\v{c}}ek and
		Vandenbussche}]{munoz2019facilitating}
	\bibinfo{author}{Mu{\~n}oz, E.}, \bibinfo{author}{Nov{\'a}{\v{c}}ek, V.},
	\bibinfo{author}{Vandenbussche, P.Y.}, \bibinfo{year}{2019}.
	\newblock \bibinfo{title}{Facilitating prediction of adverse drug reactions by
		using knowledge graphs and multi-label learning models}.
	\newblock \bibinfo{journal}{Briefings in bioinformatics} \bibinfo{volume}{20},
	\bibinfo{pages}{190--202}.
	\bibitem[{Paul et~al.(2024)Paul, Saha, Hasan, Noori and
		Moustafa}]{paul2024systematic}
	\bibinfo{author}{Paul, S.G.}, \bibinfo{author}{Saha, A.},
	\bibinfo{author}{Hasan, M.Z.}, \bibinfo{author}{Noori, S.R.H.},
	\bibinfo{author}{Moustafa, A.}, \bibinfo{year}{2024}.
	\newblock \bibinfo{title}{A systematic review of graph neural network in
		healthcare-based applications: Recent advances, trends, and future
		directions}.
	\newblock \bibinfo{journal}{IEEE access} \bibinfo{volume}{12},
	\bibinfo{pages}{15145--15170}.
	\bibitem[{Ping et~al.(2017)Ping, Watson, Han and Bui}]{ping2017individualized}
	\bibinfo{author}{Ping, P.}, \bibinfo{author}{Watson, K.}, \bibinfo{author}{Han,
		J.}, \bibinfo{author}{Bui, A.}, \bibinfo{year}{2017}.
	\newblock \bibinfo{title}{Individualized knowledge graph: a viable informatics
		path to precision medicine}.
	\newblock \bibinfo{journal}{Circulation research} \bibinfo{volume}{120},
	\bibinfo{pages}{1078--1080}.
	\bibitem[{Quan et~al.(2023)Quan, Cai, Xi, Wang and Yan}]{quan2023aimedgraph}
	\bibinfo{author}{Quan, X.}, \bibinfo{author}{Cai, W.}, \bibinfo{author}{Xi,
		C.}, \bibinfo{author}{Wang, C.}, \bibinfo{author}{Yan, L.},
	\bibinfo{year}{2023}.
	\newblock \bibinfo{title}{Aimedgraph: a comprehensive multi-relational
		knowledge graph for precision medicine}.
	\newblock \bibinfo{journal}{Database} \bibinfo{volume}{2023},
	\bibinfo{pages}{baad006}.
	\bibitem[{Renaux et~al.(2023)Renaux, Terwagne, Cochez, Tiddi, Now{\'e} and
		Lenaerts}]{renaux2023knowledge}
	\bibinfo{author}{Renaux, A.}, \bibinfo{author}{Terwagne, C.},
	\bibinfo{author}{Cochez, M.}, \bibinfo{author}{Tiddi, I.},
	\bibinfo{author}{Now{\'e}, A.}, \bibinfo{author}{Lenaerts, T.},
	\bibinfo{year}{2023}.
	\newblock \bibinfo{title}{A knowledge graph approach to predict and interpret
		disease-causing gene interactions}.
	\newblock \bibinfo{journal}{BMC bioinformatics} \bibinfo{volume}{24},
	\bibinfo{pages}{324}.
	\bibitem[{Sarani~Rad et~al.(2024)Sarani~Rad, Hendawi, Yang and
		Li}]{sarani2024personalized}
	\bibinfo{author}{Sarani~Rad, F.}, \bibinfo{author}{Hendawi, R.},
	\bibinfo{author}{Yang, X.}, \bibinfo{author}{Li, J.}, \bibinfo{year}{2024}.
	\newblock \bibinfo{title}{Personalized diabetes management with digital twins:
		a patient-centric knowledge graph approach}.
	\newblock \bibinfo{journal}{Journal of personalized medicine}
	\bibinfo{volume}{14}, \bibinfo{pages}{359}.
	\bibitem[{Shang et~al.(2021)Shang, Tian, Zhou, Zhou, Lyu, Wang, Xin, Liang, Zhu
		and Li}]{shang2021ehr}
	\bibinfo{author}{Shang, Y.}, \bibinfo{author}{Tian, Y.}, \bibinfo{author}{Zhou,
		M.}, \bibinfo{author}{Zhou, T.}, \bibinfo{author}{Lyu, K.},
	\bibinfo{author}{Wang, Z.}, \bibinfo{author}{Xin, R.},
	\bibinfo{author}{Liang, T.}, \bibinfo{author}{Zhu, S.}, \bibinfo{author}{Li,
		J.}, \bibinfo{year}{2021}.
	\newblock \bibinfo{title}{Ehr-oriented knowledge graph system: toward efficient
		utilization of non-used information buried in routine clinical practice}.
	\newblock \bibinfo{journal}{IEEE Journal of Biomedical and Health Informatics}
	\bibinfo{volume}{25}, \bibinfo{pages}{2463--2475}.
	\bibitem[{Shen et~al.(2019)Shen, Yuan, Dai, Tang, Yang and Lei}]{shen2019kgdds}
	\bibinfo{author}{Shen, Y.}, \bibinfo{author}{Yuan, K.}, \bibinfo{author}{Dai,
		J.}, \bibinfo{author}{Tang, B.}, \bibinfo{author}{Yang, M.},
	\bibinfo{author}{Lei, K.}, \bibinfo{year}{2019}.
	\newblock \bibinfo{title}{Kgdds: a system for drug-drug similarity measure in
		therapeutic substitution based on knowledge graph curation}.
	\newblock \bibinfo{journal}{Journal of medical systems} \bibinfo{volume}{43},
	\bibinfo{pages}{92}.
	\bibitem[{Sheng et~al.(2020)Sheng, Li, Bu, Dong, Zhang, Li, Li and
		Xing}]{sheng2020dsqa}
	\bibinfo{author}{Sheng, M.}, \bibinfo{author}{Li, A.}, \bibinfo{author}{Bu,
		Y.}, \bibinfo{author}{Dong, J.}, \bibinfo{author}{Zhang, Y.},
	\bibinfo{author}{Li, X.}, \bibinfo{author}{Li, C.}, \bibinfo{author}{Xing,
		C.}, \bibinfo{year}{2020}.
	\newblock \bibinfo{title}{Dsqa: a domain specific qa system for smart health
		based on knowledge graph}, in: \bibinfo{booktitle}{International conference
		on web information systems and applications},
	\bibinfo{organization}{Springer}. pp. \bibinfo{pages}{215--222}.
	\bibitem[{Shi et~al.(2017)Shi, Li, Yang, Qi, Pan and Zhou}]{shi2017semantic}
	\bibinfo{author}{Shi, L.}, \bibinfo{author}{Li, S.}, \bibinfo{author}{Yang,
		X.}, \bibinfo{author}{Qi, J.}, \bibinfo{author}{Pan, G.},
	\bibinfo{author}{Zhou, B.}, \bibinfo{year}{2017}.
	\newblock \bibinfo{title}{Semantic health knowledge graph: semantic integration
		of heterogeneous medical knowledge and services}.
	\newblock \bibinfo{journal}{BioMed research international}
	\bibinfo{volume}{2017}, \bibinfo{pages}{2858423}.
	\bibitem[{Shuai et~al.(2019)Shuai, Wei, Miao and Jin}]{shuai2019research}
	\bibinfo{author}{Shuai, Q.}, \bibinfo{author}{Wei, M.}, \bibinfo{author}{Miao,
		F.}, \bibinfo{author}{Jin, L.}, \bibinfo{year}{2019}.
	\newblock \bibinfo{title}{Research on intelligent question answering system
		based on medical knowledge graph}, in: \bibinfo{booktitle}{2019 IEEE 4th
		Advanced Information Technology, Electronic and Automation Control Conference
		(IAEAC)}, \bibinfo{organization}{IEEE}. pp. \bibinfo{pages}{240--243}.
	\bibitem[{Soman et~al.(2023)Soman, Nelson, Cerono, Goldman, Baranzini and
		Brown}]{soman2023early}
	\bibinfo{author}{Soman, K.}, \bibinfo{author}{Nelson, C.A.},
	\bibinfo{author}{Cerono, G.}, \bibinfo{author}{Goldman, S.M.},
	\bibinfo{author}{Baranzini, S.E.}, \bibinfo{author}{Brown, E.G.},
	\bibinfo{year}{2023}.
	\newblock \bibinfo{title}{Early detection of parkinson’s disease through
		enriching the electronic health record using a biomedical knowledge graph}.
	\newblock \bibinfo{journal}{Frontiers in Medicine} \bibinfo{volume}{10},
	\bibinfo{pages}{1081087}.
	\bibitem[{Soman et~al.(2024)Soman, Rose, Morris, Akbas, Smith, Peetoom,
		Villouta-Reyes, Cerono, Shi, Rizk-Jackson et~al.}]{soman2024biomedical}
	\bibinfo{author}{Soman, K.}, \bibinfo{author}{Rose, P.W.},
	\bibinfo{author}{Morris, J.H.}, \bibinfo{author}{Akbas, R.E.},
	\bibinfo{author}{Smith, B.}, \bibinfo{author}{Peetoom, B.},
	\bibinfo{author}{Villouta-Reyes, C.}, \bibinfo{author}{Cerono, G.},
	\bibinfo{author}{Shi, Y.}, \bibinfo{author}{Rizk-Jackson, A.}, et~al.,
	\bibinfo{year}{2024}.
	\newblock \bibinfo{title}{Biomedical knowledge graph-optimized prompt
		generation for large language models}.
	\newblock \bibinfo{journal}{Bioinformatics} \bibinfo{volume}{40},
	\bibinfo{pages}{btae560}.
	\bibitem[{Su et~al.(2022a)Su, Dougherty, Jiang and Jin}]{su2022interactive}
	\bibinfo{author}{Su, J.}, \bibinfo{author}{Dougherty, E.T.},
	\bibinfo{author}{Jiang, S.}, \bibinfo{author}{Jin, F.},
	\bibinfo{year}{2022}a.
	\newblock \bibinfo{title}{An interactive knowledge graph based platform for
		covid-19 clinical research}, in: \bibinfo{booktitle}{Proceedings of the
		Fifteenth ACM International Conference on Web Search and Data Mining}, pp.
	\bibinfo{pages}{1609--1612}.
	\bibitem[{Su et~al.(2022b)Su, Hu, You, Hu and Zhao}]{su2022attention}
	\bibinfo{author}{Su, X.}, \bibinfo{author}{Hu, L.}, \bibinfo{author}{You, Z.},
	\bibinfo{author}{Hu, P.}, \bibinfo{author}{Zhao, B.}, \bibinfo{year}{2022}b.
	\newblock \bibinfo{title}{Attention-based knowledge graph representation
		learning for predicting drug-drug interactions}.
	\newblock \bibinfo{journal}{Briefings in bioinformatics} \bibinfo{volume}{23},
	\bibinfo{pages}{bbac140}.
	\bibitem[{Sun et~al.(2020)Sun, Xiao, Zhu, He, Zhang, Xu, Hou, Li, Ni and
		Xie}]{sun2020medical}
	\bibinfo{author}{Sun, H.}, \bibinfo{author}{Xiao, J.}, \bibinfo{author}{Zhu,
		W.}, \bibinfo{author}{He, Y.}, \bibinfo{author}{Zhang, S.},
	\bibinfo{author}{Xu, X.}, \bibinfo{author}{Hou, L.}, \bibinfo{author}{Li,
		J.}, \bibinfo{author}{Ni, Y.}, \bibinfo{author}{Xie, G.},
	\bibinfo{year}{2020}.
	\newblock \bibinfo{title}{Medical knowledge graph to enhance fraud, waste, and
		abuse detection on claim data: model development and performance evaluation}.
	\newblock \bibinfo{journal}{JMIR Medical Informatics} \bibinfo{volume}{8},
	\bibinfo{pages}{e17653}.
	\bibitem[{Sun et~al.(2021a)Sun, Huang, Li, Xu and Hou}]{sun2021kgbref}
	\bibinfo{author}{Sun, Y.}, \bibinfo{author}{Huang, Z.}, \bibinfo{author}{Li,
		J.}, \bibinfo{author}{Xu, Z.}, \bibinfo{author}{Hou, L.},
	\bibinfo{year}{2021}a.
	\newblock \bibinfo{title}{Kgbref: A knowledge graph based biomedical relation
		extraction framework}, in: \bibinfo{booktitle}{Proceedings of the 2nd
		International Symposium on Artificial Intelligence for Medicine Sciences},
	pp. \bibinfo{pages}{114--119}.
	\bibitem[{Sun et~al.(2021b)Sun, Li, Zheng, Zhu and Wu}]{sun2021intelligent}
	\bibinfo{author}{Sun, Y.}, \bibinfo{author}{Li, Y.}, \bibinfo{author}{Zheng,
		B.}, \bibinfo{author}{Zhu, S.}, \bibinfo{author}{Wu, M.},
	\bibinfo{year}{2021}b.
	\newblock \bibinfo{title}{An intelligent question-answering system for myopia
		prevention and control based on knowledge graph}, in:
	\bibinfo{booktitle}{Proceedings of the 2nd International Symposium on
		Artificial Intelligence for Medicine Sciences}, pp. \bibinfo{pages}{80--87}.
	\bibitem[{Tao et~al.(2020)Tao, Pham, Zhang, Yong, Goh, Zhang and
		Cai}]{tao2020mining}
	\bibinfo{author}{Tao, X.}, \bibinfo{author}{Pham, T.}, \bibinfo{author}{Zhang,
		J.}, \bibinfo{author}{Yong, J.}, \bibinfo{author}{Goh, W.P.},
	\bibinfo{author}{Zhang, W.}, \bibinfo{author}{Cai, Y.}, \bibinfo{year}{2020}.
	\newblock \bibinfo{title}{Mining health knowledge graph for health risk
		prediction}.
	\newblock \bibinfo{journal}{World Wide Web} \bibinfo{volume}{23},
	\bibinfo{pages}{2341--2362}.
	\bibitem[{Thukral et~al.(2023)Thukral, Dhiman, Meher and
		Bedi}]{thukral2023knowledge}
	\bibinfo{author}{Thukral, A.}, \bibinfo{author}{Dhiman, S.},
	\bibinfo{author}{Meher, R.}, \bibinfo{author}{Bedi, P.},
	\bibinfo{year}{2023}.
	\newblock \bibinfo{title}{Knowledge graph enrichment from clinical narratives
		using nlp, ner, and biomedical ontologies for healthcare applications}.
	\newblock \bibinfo{journal}{International Journal of Information Technology}
	\bibinfo{volume}{15}, \bibinfo{pages}{53--65}.
	\bibitem[{Tran et~al.(2021)Tran, Felfernig, Trattner and
		Holzinger}]{tran2021recommender}
	\bibinfo{author}{Tran, T.N.T.}, \bibinfo{author}{Felfernig, A.},
	\bibinfo{author}{Trattner, C.}, \bibinfo{author}{Holzinger, A.},
	\bibinfo{year}{2021}.
	\newblock \bibinfo{title}{Recommender systems in the healthcare domain:
		state-of-the-art and research issues}.
	\newblock \bibinfo{journal}{Journal of Intelligent Information Systems}
	\bibinfo{volume}{57}, \bibinfo{pages}{171--201}.
	\bibitem[{Varshney et~al.(2023)Varshney, Zafar, Behera and
		Ekbal}]{varshney2023knowledge}
	\bibinfo{author}{Varshney, D.}, \bibinfo{author}{Zafar, A.},
	\bibinfo{author}{Behera, N.K.}, \bibinfo{author}{Ekbal, A.},
	\bibinfo{year}{2023}.
	\newblock \bibinfo{title}{Knowledge graph assisted end-to-end medical dialog
		generation}.
	\newblock \bibinfo{journal}{Artificial Intelligence in Medicine}
	\bibinfo{volume}{139}, \bibinfo{pages}{102535}.
	\bibitem[{Vidal et~al.(2025)Vidal, Chudasama, Huang, Purohit and
		Torrente}]{vidal2025integrating}
	\bibinfo{author}{Vidal, M.E.}, \bibinfo{author}{Chudasama, Y.},
	\bibinfo{author}{Huang, H.}, \bibinfo{author}{Purohit, D.},
	\bibinfo{author}{Torrente, M.}, \bibinfo{year}{2025}.
	\newblock \bibinfo{title}{Integrating knowledge graphs with symbolic ai: The
		path to interpretable hybrid ai systems in medicine}.
	\newblock \bibinfo{journal}{Journal of Web Semantics} \bibinfo{volume}{84},
	\bibinfo{pages}{100856}.
	\bibitem[{Vilela et~al.(2023)Vilela, Asif, Marques, Santos, Rasga, Vicente and
		Martiniano}]{vilela2023biomedical}
	\bibinfo{author}{Vilela, J.}, \bibinfo{author}{Asif, M.},
	\bibinfo{author}{Marques, A.R.}, \bibinfo{author}{Santos, J.X.},
	\bibinfo{author}{Rasga, C.}, \bibinfo{author}{Vicente, A.},
	\bibinfo{author}{Martiniano, H.}, \bibinfo{year}{2023}.
	\newblock \bibinfo{title}{Biomedical knowledge graph embeddings for
		personalized medicine: Predicting disease-gene associations}.
	\newblock \bibinfo{journal}{Expert Systems} \bibinfo{volume}{40},
	\bibinfo{pages}{e13181}.
	\bibitem[{Wang et~al.(2019)Wang, Zhao, Xie, Li and Guo}]{wang2019knowledge}
	\bibinfo{author}{Wang, H.}, \bibinfo{author}{Zhao, M.}, \bibinfo{author}{Xie,
		X.}, \bibinfo{author}{Li, W.}, \bibinfo{author}{Guo, M.},
	\bibinfo{year}{2019}.
	\newblock \bibinfo{title}{Knowledge graph convolutional networks for
		recommender systems}, in: \bibinfo{booktitle}{The world wide web conference},
	pp. \bibinfo{pages}{3307--3313}.
	\bibitem[{Wang et~al.(2021)Wang, Ma, Si, Tang, Wang, Li, Ouyang, Gong, Tang, He
		et~al.}]{wang2021adverse}
	\bibinfo{author}{Wang, M.}, \bibinfo{author}{Ma, X.}, \bibinfo{author}{Si, J.},
	\bibinfo{author}{Tang, H.}, \bibinfo{author}{Wang, H.}, \bibinfo{author}{Li,
		T.}, \bibinfo{author}{Ouyang, W.}, \bibinfo{author}{Gong, L.},
	\bibinfo{author}{Tang, Y.}, \bibinfo{author}{He, X.}, et~al.,
	\bibinfo{year}{2021}.
	\newblock \bibinfo{title}{Adverse drug reaction discovery using a
		tumor-biomarker knowledge graph}.
	\newblock \bibinfo{journal}{Frontiers in genetics} \bibinfo{volume}{11},
	\bibinfo{pages}{625659}.
	\bibitem[{Wang(2020)}]{wang2020novel}
	\bibinfo{author}{Wang, Y.}, \bibinfo{year}{2020}.
	\newblock \bibinfo{title}{A novel chinese traditional medicine prescription
		recommendation system based on knowledge graph}, in:
	\bibinfo{booktitle}{Journal of Physics: Conference Series},
	\bibinfo{organization}{IOP Publishing}. p. \bibinfo{pages}{012019}.
	\bibitem[{Wu et~al.(2023a)Wu, He, Mao, Li and Cambria}]{wu2023megacare}
	\bibinfo{author}{Wu, J.}, \bibinfo{author}{He, K.}, \bibinfo{author}{Mao, R.},
	\bibinfo{author}{Li, C.}, \bibinfo{author}{Cambria, E.},
	\bibinfo{year}{2023}a.
	\newblock \bibinfo{title}{Megacare: Knowledge-guided multi-view hypergraph
		predictive framework for healthcare}.
	\newblock \bibinfo{journal}{Information Fusion} \bibinfo{volume}{100},
	\bibinfo{pages}{101939}.
	\bibitem[{Wu et~al.(2023b)Wu, Duan, Pan and Li}]{wu2023medical}
	\bibinfo{author}{Wu, X.}, \bibinfo{author}{Duan, J.}, \bibinfo{author}{Pan,
		Y.}, \bibinfo{author}{Li, M.}, \bibinfo{year}{2023}b.
	\newblock \bibinfo{title}{Medical knowledge graph: Data sources, construction,
		reasoning, and applications}.
	\newblock \bibinfo{journal}{Big data mining and analytics} \bibinfo{volume}{6},
	\bibinfo{pages}{201--217}.
	\bibitem[{Xu et~al.(2023)Xu, Chu, Yang, Wang, Zou, Ding, Zhao, Wang and
		Xie}]{xu2023seqcare}
	\bibinfo{author}{Xu, Y.}, \bibinfo{author}{Chu, X.}, \bibinfo{author}{Yang,
		K.}, \bibinfo{author}{Wang, Z.}, \bibinfo{author}{Zou, P.},
	\bibinfo{author}{Ding, H.}, \bibinfo{author}{Zhao, J.},
	\bibinfo{author}{Wang, Y.}, \bibinfo{author}{Xie, B.}, \bibinfo{year}{2023}.
	\newblock \bibinfo{title}{Seqcare: Sequential training with external medical
		knowledge graph for diagnosis prediction in healthcare data}, in:
	\bibinfo{booktitle}{Proceedings of the ACM Web Conference 2023}, pp.
	\bibinfo{pages}{2819--2830}.
	\bibitem[{Xu et~al.(2024)Xu, Gu, Xu, Topaz, Guo, Kang, Sun and
		Li}]{xu2024developing}
	\bibinfo{author}{Xu, Z.}, \bibinfo{author}{Gu, Y.}, \bibinfo{author}{Xu, X.},
	\bibinfo{author}{Topaz, M.}, \bibinfo{author}{Guo, Z.},
	\bibinfo{author}{Kang, H.}, \bibinfo{author}{Sun, L.}, \bibinfo{author}{Li,
		J.}, \bibinfo{year}{2024}.
	\newblock \bibinfo{title}{Developing a personalized meal recommendation system
		for chinese older adults: observational cohort study}.
	\newblock \bibinfo{journal}{JMIR Formative Research} \bibinfo{volume}{8},
	\bibinfo{pages}{e52170}.
	\bibitem[{Yang et~al.(2022a)Yang, Gessner, Duerksen, Biber, Binder, Ozturk,
		Foote, McEntire, Stirling, Ding et~al.}]{yang2022knowledge}
	\bibinfo{author}{Yang, J.J.}, \bibinfo{author}{Gessner, C.R.},
	\bibinfo{author}{Duerksen, J.L.}, \bibinfo{author}{Biber, D.},
	\bibinfo{author}{Binder, J.L.}, \bibinfo{author}{Ozturk, M.},
	\bibinfo{author}{Foote, B.}, \bibinfo{author}{McEntire, R.},
	\bibinfo{author}{Stirling, K.}, \bibinfo{author}{Ding, Y.}, et~al.,
	\bibinfo{year}{2022}a.
	\newblock \bibinfo{title}{Knowledge graph analytics platform with lincs and idg
		for parkinson's disease target illumination}.
	\newblock \bibinfo{journal}{BMC bioinformatics} \bibinfo{volume}{23},
	\bibinfo{pages}{37}.
	\bibitem[{Yang et~al.(2023a)Yang, Xu, Zou, Ding, Zhao, Wang and
		Xie}]{yang2023kerprint}
	\bibinfo{author}{Yang, K.}, \bibinfo{author}{Xu, Y.}, \bibinfo{author}{Zou,
		P.}, \bibinfo{author}{Ding, H.}, \bibinfo{author}{Zhao, J.},
	\bibinfo{author}{Wang, Y.}, \bibinfo{author}{Xie, B.}, \bibinfo{year}{2023}a.
	\newblock \bibinfo{title}{Kerprint: local-global knowledge graph enhanced
		diagnosis prediction for retrospective and prospective interpretations}, in:
	\bibinfo{booktitle}{Proceedings of the AAAI Conference on Artificial
		Intelligence}, pp. \bibinfo{pages}{5357--5365}.
	\bibitem[{Yang et~al.(2022b)Yang, Ye, Cheng, Zhang, Lan and
		Zou}]{yang2022decision}
	\bibinfo{author}{Yang, R.}, \bibinfo{author}{Ye, Q.}, \bibinfo{author}{Cheng,
		C.}, \bibinfo{author}{Zhang, S.}, \bibinfo{author}{Lan, Y.},
	\bibinfo{author}{Zou, J.}, \bibinfo{year}{2022}b.
	\newblock \bibinfo{title}{Decision-making system for the diagnosis of syndrome
		based on traditional chinese medicine knowledge graph}.
	\newblock \bibinfo{journal}{Evidence-Based Complementary and Alternative
		Medicine} \bibinfo{volume}{2022}, \bibinfo{pages}{8693937}.
	\bibitem[{Yang et~al.(2023b)Yang, Lu and Yan}]{yang2023comprehensive}
	\bibinfo{author}{Yang, Y.}, \bibinfo{author}{Lu, Y.}, \bibinfo{author}{Yan,
		W.}, \bibinfo{year}{2023}b.
	\newblock \bibinfo{title}{A comprehensive review on knowledge graphs for
		complex diseases}.
	\newblock \bibinfo{journal}{Briefings in bioinformatics} \bibinfo{volume}{24},
	\bibinfo{pages}{bbac543}.
	\bibitem[{Ye et~al.(2021)Ye, Hsieh, Yang, Kang, Chen, Cao, He and
		Hou}]{ye2021unified}
	\bibinfo{author}{Ye, Q.}, \bibinfo{author}{Hsieh, C.Y.}, \bibinfo{author}{Yang,
		Z.}, \bibinfo{author}{Kang, Y.}, \bibinfo{author}{Chen, J.},
	\bibinfo{author}{Cao, D.}, \bibinfo{author}{He, S.}, \bibinfo{author}{Hou,
		T.}, \bibinfo{year}{2021}.
	\newblock \bibinfo{title}{A unified drug--target interaction prediction
		framework based on knowledge graph and recommendation system}.
	\newblock \bibinfo{journal}{Nature communications} \bibinfo{volume}{12},
	\bibinfo{pages}{6775}.
	\bibitem[{Yin et~al.(2022)Yin, Zhang, Wang, Wang, Zhang and
		Li}]{yin2022question}
	\bibinfo{author}{Yin, Y.}, \bibinfo{author}{Zhang, L.}, \bibinfo{author}{Wang,
		Y.}, \bibinfo{author}{Wang, M.}, \bibinfo{author}{Zhang, Q.},
	\bibinfo{author}{Li, G.z.}, \bibinfo{year}{2022}.
	\newblock \bibinfo{title}{Question answering system based on knowledge graph in
		traditional chinese medicine diagnosis and treatment of viral hepatitis b}.
	\newblock \bibinfo{journal}{BioMed research international}
	\bibinfo{volume}{2022}, \bibinfo{pages}{7139904}.
	\bibitem[{Yu et~al.(2022)Yu, Tabatabaei, Mezei, Zhong, Chen, Li, Li, Shu and
		Shu}]{yu2022improving}
	\bibinfo{author}{Yu, G.}, \bibinfo{author}{Tabatabaei, M.},
	\bibinfo{author}{Mezei, J.}, \bibinfo{author}{Zhong, Q.},
	\bibinfo{author}{Chen, S.}, \bibinfo{author}{Li, Z.}, \bibinfo{author}{Li,
		J.}, \bibinfo{author}{Shu, L.}, \bibinfo{author}{Shu, Q.},
	\bibinfo{year}{2022}.
	\newblock \bibinfo{title}{Improving chronic disease management for children
		with knowledge graphs and artificial intelligence}.
	\newblock \bibinfo{journal}{Expert Systems with Applications}
	\bibinfo{volume}{201}, \bibinfo{pages}{117026}.
	\bibitem[{Zeng and Nakano(2020)}]{zeng2020exploiting}
	\bibinfo{author}{Zeng, J.}, \bibinfo{author}{Nakano, Y.I.},
	\bibinfo{year}{2020}.
	\newblock \bibinfo{title}{Exploiting a large-scale knowledge graph for question
		generation in food preference interview systems}, in:
	\bibinfo{booktitle}{Companion Proceedings of the 25th International
		Conference on Intelligent User Interfaces}, pp. \bibinfo{pages}{53--54}.
	\bibitem[{Zeng et~al.(2022)Zeng, Tu, Liu, Fu and Su}]{zeng2022toward}
	\bibinfo{author}{Zeng, X.}, \bibinfo{author}{Tu, X.}, \bibinfo{author}{Liu,
		Y.}, \bibinfo{author}{Fu, X.}, \bibinfo{author}{Su, Y.},
	\bibinfo{year}{2022}.
	\newblock \bibinfo{title}{Toward better drug discovery with knowledge graph}.
	\newblock \bibinfo{journal}{Current opinion in structural biology}
	\bibinfo{volume}{72}, \bibinfo{pages}{114--126}.
	\bibitem[{Zhang et~al.(2021a)Zhang, Sun, Diao, Zhao and
		Shu}]{zhang2021prediction}
	\bibinfo{author}{Zhang, F.}, \bibinfo{author}{Sun, B.}, \bibinfo{author}{Diao,
		X.}, \bibinfo{author}{Zhao, W.}, \bibinfo{author}{Shu, T.},
	\bibinfo{year}{2021}a.
	\newblock \bibinfo{title}{Prediction of adverse drug reactions based on
		knowledge graph embedding}.
	\newblock \bibinfo{journal}{BMC medical informatics and decision making}
	\bibinfo{volume}{21}, \bibinfo{pages}{38}.
	\bibitem[{Zhang(2021)}]{zhang2021knowledge}
	\bibinfo{author}{Zhang, K.}, \bibinfo{year}{2021}.
	\newblock \bibinfo{title}{A knowledge graph based medical intelligent question
		answering system}, in: \bibinfo{booktitle}{2021 IEEE International Conference
		on Computer Science, Electronic Information Engineering and Intelligent
		Control Technology (CEI)}, \bibinfo{organization}{IEEE}. pp.
	\bibinfo{pages}{460--465}.
	\bibitem[{Zhang et~al.(2021b)Zhang, Lin and Zhang}]{zhang2021discovering}
	\bibinfo{author}{Zhang, S.}, \bibinfo{author}{Lin, X.}, \bibinfo{author}{Zhang,
		X.}, \bibinfo{year}{2021}b.
	\newblock \bibinfo{title}{Discovering dti and ddi by knowledge graph with mhrw
		and improved neural network}, in: \bibinfo{booktitle}{2021 IEEE International
		Conference on Bioinformatics and Biomedicine (BIBM)},
	\bibinfo{organization}{IEEE}. pp. \bibinfo{pages}{588--593}.
	\bibitem[{Zhang et~al.(2020)Zhang, Sheng, Zhou, Wang, Han, Zhang, Xing and
		Dong}]{zhang2020hkgb}
	\bibinfo{author}{Zhang, Y.}, \bibinfo{author}{Sheng, M.},
	\bibinfo{author}{Zhou, R.}, \bibinfo{author}{Wang, Y.}, \bibinfo{author}{Han,
		G.}, \bibinfo{author}{Zhang, H.}, \bibinfo{author}{Xing, C.},
	\bibinfo{author}{Dong, J.}, \bibinfo{year}{2020}.
	\newblock \bibinfo{title}{Hkgb: an inclusive, extensible, intelligent,
		semi-auto-constructed knowledge graph framework for healthcare with
		clinicians’ expertise incorporated}.
	\newblock \bibinfo{journal}{Information Processing \& Management}
	\bibinfo{volume}{57}, \bibinfo{pages}{102324}.
	\bibitem[{Zhao et~al.(2023)Zhao, Qi, Chen, Qiao, Bu, Wu, Luo, Wang, Zhang and
		Zhao}]{zhao2023biological}
	\bibinfo{author}{Zhao, L.}, \bibinfo{author}{Qi, X.}, \bibinfo{author}{Chen,
		Y.}, \bibinfo{author}{Qiao, Y.}, \bibinfo{author}{Bu, D.},
	\bibinfo{author}{Wu, Y.}, \bibinfo{author}{Luo, Y.}, \bibinfo{author}{Wang,
		S.}, \bibinfo{author}{Zhang, R.}, \bibinfo{author}{Zhao, Y.},
	\bibinfo{year}{2023}.
	\newblock \bibinfo{title}{Biological knowledge graph-guided investigation of
		immune therapy response in cancer with graph neural network}.
	\newblock \bibinfo{journal}{Briefings in Bioinformatics} \bibinfo{volume}{24},
	\bibinfo{pages}{bbad023}.
	\bibitem[{Zheng et~al.(2021)Zheng, Rao, Song, Zhang, Xiao, Fang, Yang and
		Niu}]{zheng2021pharmkg}
	\bibinfo{author}{Zheng, S.}, \bibinfo{author}{Rao, J.}, \bibinfo{author}{Song,
		Y.}, \bibinfo{author}{Zhang, J.}, \bibinfo{author}{Xiao, X.},
	\bibinfo{author}{Fang, E.F.}, \bibinfo{author}{Yang, Y.},
	\bibinfo{author}{Niu, Z.}, \bibinfo{year}{2021}.
	\newblock \bibinfo{title}{Pharmkg: a dedicated knowledge graph benchmark for
		biomedical data mining}.
	\newblock \bibinfo{journal}{Briefings in bioinformatics} \bibinfo{volume}{22},
	\bibinfo{pages}{bbaa344}.
	\bibitem[{Zhou et~al.(2023)Zhou, E, Kuang, Tan, Xie, Li and
		Luo}]{zhou2023corrigendum}
	\bibinfo{author}{Zhou, G.}, \bibinfo{author}{E, H.}, \bibinfo{author}{Kuang,
		Z.}, \bibinfo{author}{Tan, L.}, \bibinfo{author}{Xie, X.},
	\bibinfo{author}{Li, J.}, \bibinfo{author}{Luo, H.}, \bibinfo{year}{2023}.
	\newblock \bibinfo{title}{Corrigendum to clinical decision support system for
		hypertension medication based on knowledge graph::[computer methods and
		programs in biomedicine volume 227 (2022)/107220]} .
	\bibitem[{Zhu et~al.(2020)Zhu, Che, Jin, Zhang, Su and Wang}]{zhu2020knowledge}
	\bibinfo{author}{Zhu, Y.}, \bibinfo{author}{Che, C.}, \bibinfo{author}{Jin,
		B.}, \bibinfo{author}{Zhang, N.}, \bibinfo{author}{Su, C.},
	\bibinfo{author}{Wang, F.}, \bibinfo{year}{2020}.
	\newblock \bibinfo{title}{Knowledge-driven drug repurposing using a
		comprehensive drug knowledge graph}.
	\newblock \bibinfo{journal}{Health Informatics Journal} \bibinfo{volume}{26},
	\bibinfo{pages}{2737--2750}.
	\bibitem[{Zhuang et~al.(2025)Zhuang, Yu, Jiang and Ge}]{zhuang2025tcm}
	\bibinfo{author}{Zhuang, Y.}, \bibinfo{author}{Yu, L.}, \bibinfo{author}{Jiang,
		N.}, \bibinfo{author}{Ge, Y.}, \bibinfo{year}{2025}.
	\newblock \bibinfo{title}{Tcm-kllama: Intelligent generation model for
		traditional chinese medicine prescriptions based on knowledge graph and large
		language model}.
	\newblock \bibinfo{journal}{Computers in Biology and Medicine}
	\bibinfo{volume}{189}, \bibinfo{pages}{109887}.
	
\end{thebibliography}



\end{document}